\documentclass[10pt,twocolumn,letterpaper]{article}

\usepackage[pagenumbers]{cvpr} 

\usepackage{graphicx}
\usepackage{amsmath}
\usepackage{amssymb}
\usepackage{booktabs}
\usepackage{bm}
\usepackage{multirow}
\usepackage{multicol}
\usepackage{adjustbox}
\usepackage[table,xcdraw,dvipsnames]{xcolor}
\usepackage{nicefrac}       
\usepackage{paralist}
\usepackage{mathtools}

\usepackage[pagebackref=true,breaklinks=true,colorlinks,bookmarks=false,citecolor=blue,linkcolor=blue]{hyperref}

\usepackage[capitalize]{cleveref}
\crefname{section}{Sec.}{Secs.}
\Crefname{section}{Section}{Sections}
\Crefname{table}{Table}{Tables}
\crefname{table}{Tab.}{Tabs.}

\newcommand\bluecirc{{\color{ProcessBlue}\bullet}\mathllap{\circ}}
\newcommand\redcirc{{\color{Red}\bullet}\mathllap{\circ}}
\def\newin{\!\in\!}
\def\neweq{\!=\!}

\definecolor{mygreen}{rgb}{0.4, 0.69, 0.2}

\newcommand*{\affaddr}[1]{\small #1} 
\newcommand*{\affmark}[1][*]{\textsuperscript{#1}}

\title{Spatio-temporal Relation Modeling for Few-shot Action Recognition}
\begin{document}

\author{
Anirudh Thatipelli\affmark[1] \quad
Sanath Narayan\affmark[2] \quad
Salman Khan\affmark[1,4]  \\
Rao Muhammad Anwer\affmark[1,3] \quad
Fahad Shahbaz Khan\affmark[1,5] \quad 
Bernard Ghanem\affmark[6]\\
\affaddr{\affmark[1]Mohamed Bin Zayed University of Artificial Intelligence} \quad
\affaddr{\affmark[2]Inception Institute of Artificial Intelligence} \quad
\affaddr{\affmark[3]Aalto University}
\\
\affaddr{\affmark[4]Australian National University} \quad
\affaddr{\affmark[5]CVL, Linköping University} \quad
\affaddr{\affmark[6]King Abdullah University of Science \& Technology}
}

\maketitle

\begin{abstract}
We propose a novel few-shot action recognition framework, STRM, which enhances class-specific feature discriminability while simultaneously learning higher-order temporal representations. The focus of our approach is a novel spatio-temporal enrichment module that aggregates spatial and temporal contexts with dedicated local patch-level and global frame-level feature enrichment sub-modules. Local patch-level enrichment captures the appearance-based characteristics of actions.
On the other hand, global frame-level enrichment explicitly encodes the broad temporal context, thereby capturing the relevant object features over time. The resulting spatio-temporally enriched representations are then utilized to learn the relational matching between query and support action sub-sequences. We further introduce a query-class similarity classifier on the patch-level enriched features to enhance class-specific feature discriminability by reinforcing the feature learning at different stages in the proposed framework.
Experiments are performed on four few-shot action recognition benchmarks: Kinetics, SSv2, HMDB51 and UCF101. Our extensive ablation study reveals the benefits of the proposed contributions. Furthermore, our approach sets a new state-of-the-art on all four benchmarks. On the challenging SSv2 benchmark, our approach achieves an absolute gain of $3.5$\% in classification accuracy, as compared to the best existing method in the literature. Our code and models are available at \url{https://github.com/Anirudh257/strm}.
\end{abstract}

\section{Introduction}
Few-shot (FS) action recognition is a challenging computer vision problem, where the task is to classify an unlabelled query video into one of the action categories in the support set having limited samples per action class. The problem setting is particularly relevant for fine-grained action recognition~\cite{Goyal_2017_ICCV}, since it is challenging to collect sufficient labelled examples~\cite{carreira2017quo,damen2021rescaling}. 
Most existing FS action recognition methods typically search for either a single support video ~\cite{arn} or an average representation of a support class ~\cite{tarn,otam}. However, these approaches utilize only frame-level representations and do not explicitly exploit video sub-sequences for temporal relationship modeling.

In the context of FS action recognition, modeling temporal relationships between a query video and limited support actions is a major challenge, since actions are typically performed at various speeds and occur at different time instants (temporal offsets). Further, video representations are desired to encode the relevant information from multiple sub-actions that constitute an action for enhanced matching between query and support videos. Moreover, an effective representation of spatial and temporal contexts of actions is crucial to distinguish fine-grained classes requiring temporal relational reasoning, where actions can be performed with different objects in various backgrounds, \eg, \textit{spilling something behind something}.

\begin{figure*}[t]
    \centering
    \includegraphics[clip=true, trim=0em 0em 0em 0em, width=1\textwidth]{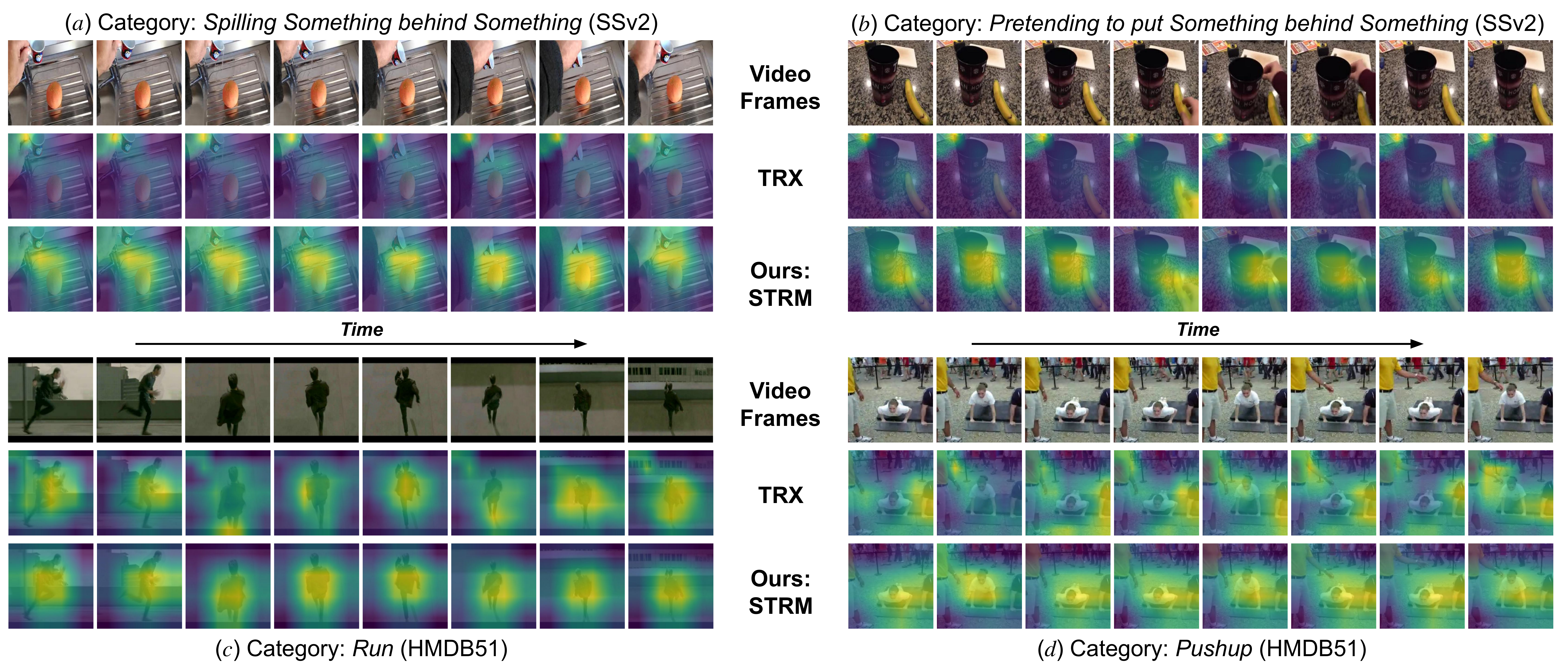}\vspace{-0.2cm}
    \caption{\textbf{Example attention map visualizations} obtained from the recently introduced TRX~\cite{trx} and our proposed STRM approach on four examples from the SSV2 and HMDB51 test set. The attention maps measure the activation magnitude of latent features. TRX struggles in the case of spatial and temporal context variations that are commonly encountered in actions performed with different objects and backgrounds, \eg, 5$^{th}$ and 6$^{th}$ frame from the left in (\textit{b}), where the regions corresponding to actions are not emphasized. Similarly, while background region is also emphasized in 3$^{rd}$ and 6$^{th}$ frame from the left in (c), the action in the 2$^{nd}$ and 3$^{rd}$ frame from the left in (\textit{d}) is not accurately captured due to the distractor motion from the moving hand of another person.  Our STRM approach explicitly enhances class-specific feature discriminability through spatio-temporal context aggregation and intermediate latent feature classification. This leads to better matching between query and limited support action instances. Additional examples are presented in Fig.~\ref{fig:method_viz} and Sec.~\ref{sec:qual}. \vspace{-0.3cm}
}
    \label{fig:intro_fig}
\end{figure*}

The aforementioned problem of temporal relationship modeling is recently explored by Temporal-Relational CrossTransformers (TRX) \cite{trx}, which compares the sub-sequences of query and support videos in a part-based manner to tackle the issue of varying speed and offsets of actions. Additionally, TRX models complex higher-order temporal relations by representing sub-sequences as tuples with different cardinalities. However, TRX struggles in the case of actions performed with different objects and background (see Fig.~\ref{fig:intro_fig}). This is likely due to not explicitly utilizing the available rich spatio-temporal contextual information during temporal relationship modeling. Furthermore, the tuple representations in TRX are fixed requiring a separate CrossTransformer~\cite{crosstransformers} branch per cardinality, which affects the model flexibility. Here, we set out to collectively address the above issues while modeling temporal relationships between query and limited support actions.

In this work, we argue that both local patch features in a frame and global frame features in a video are desirable cues to effectively enrich the encoding of spatial as well as temporal contextual information. Such feature enrichment improves class-specific discriminability, enabling focus on relevant objects and their corresponding motion in a video. In addition, learning to classify feature representations at different stages is expected to reinforce the model to look for class-separable features, thereby further improving the class-specific discriminability. Moreover, this class-specific discriminability is attainable through a reduced set of cardinalities generated by the automatic learning of higher-order temporal relationships.

\noindent\textbf{Contributions:}
We introduce an FS action recognition framework that comprises spatio-temporal enrichment and temporal relationship modeling modules along with a query-class similarity classifier. The spatio-temporal enrichment module comprises local patch-level enrichment (PLE) and global frame-level enrichment (FLE) sub-modules. The PLE enriches local patch features with spatial context by attending to all patches in a frame, in a sample-dependent manner, in order to capture the appearance-based similarities as well as dissimilarities among the action categories. On the other hand, the FLE enriches global frame features with temporal context by persistent relationship memory-based (sample-agnostic) aggregation that encompasses the entire receptive field in order to capture the relevant object motion in a video. 
The resulting enriched frame-level global representations are then utilized in the temporal relationship modeling (TRM) module to learn the temporal relations between query and support actions. 
Our TRM module does not rely on multiple cardinalities to model higher-order relations. Instead, it utilizes the spatio-temporal enrichment module to learn higher-order temporal representations at lower cardinalities.
Moreover, we introduce a query-class similarity classifier that further enhances class-specific discriminability of the spatio-temporally enriched features by learning to classify representations from intermediate layer outputs. 

We conduct extensive experiments on four FS action recognition benchmarks: Kinetics \cite{carreira2017quo}, SSv2 \cite{Goyal_2017_ICCV}, HMDB51, \cite{hmdb} and UCF101 \cite{ucf101}. Our extensive ablations show that both the proposed spatio-temporal enrichment and query-class similarity classifier enhance feature discriminability, leading to significant improvements over the baseline. The spatio-temporal enrichment module further enables the modeling of temporal relationships using a single cardinality. Our approach outperforms existing FS action recognition methods in the literature on all four benchmarks. On the challenging SSv2 benchmark, our approach achieves classification accuracy of $68.1\%$ with an absolute gain of $3.5\%$ over the recently introduced TRX~\cite{trx}, when employing the ResNet-50 backbone. Fig.~\ref{fig:intro_fig} shows a comparison of our approach with TRX, in terms of attention map visualizations, on examples from SSv2 and HMDB51.

\section{Preliminaries}
\noindent\textbf{Problem Formulation:} The goal of few-shot (FS) action recognition is to classify an unlabelled query video into one of the $C$ action classes in the `support set' comprising $K$ labelled instances for each class that is unseen during training. To this end, let $Q=\{q_1,\cdots,q_L\}$ denote a query video of $L$ frames that is to be classified into a class $c\in C$. Moreover, let $S^c$ be the support set of $K$ videos for an action class $c$ with the $k^{th}$ video denoted as $S_k^c=\{s_{k1}^c,\cdots,s_{kL}^c\}$. For simplicity, we represent each video as a sequence of uniformly sampled $L$ frames. In this work, we follow an episodic training paradigm as in~\cite{laenen2020episodes}, where few-shot tasks are randomly sampled from the training set for learning the $C$-way $K$-shot classification task in each episode. 
Next, we describe the baseline FS action recognition framework.

\begin{figure*}[t!]
  \centering
    \includegraphics[width=0.91\textwidth]{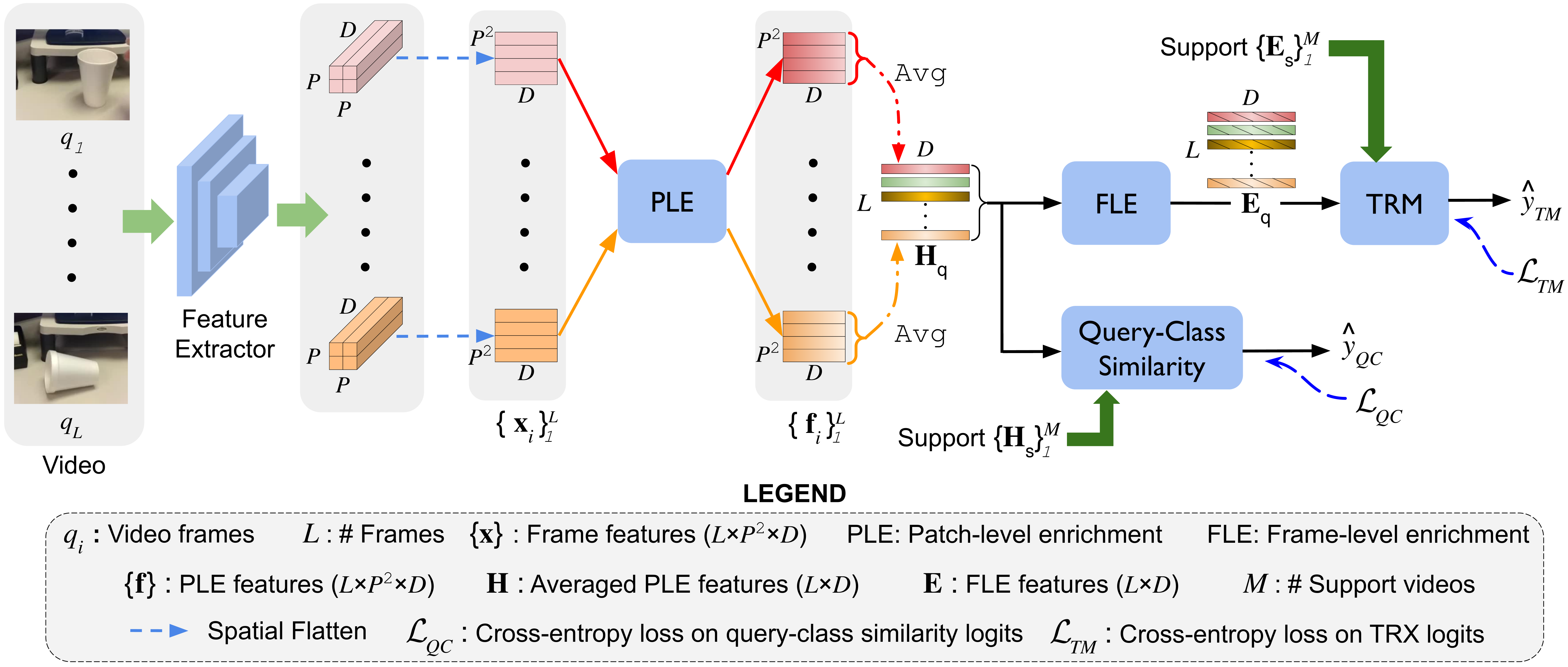}\vspace{-0.35cm}
    \caption{\textbf{Proposed STRM architecture} (Sec.~\ref{sec:overall_arch}). Spatially flattened $D$-dimensional features $\mathbf{x}_i \newin \mathbb{R}^{P^2\times D}$ are extracted for video frames $q_i$ ($i \newin [1,L]$). Here, $P^2$ is the number of patches. The features $\mathbf{x}_i$ are input to a patch-level enrichment (PLE, Sec.~\ref{sec:PLE}) block, which attends to the spatial context across patches in a frame and outputs spatially enriched features $\mathbf{f}_i \newin \mathbb{R}^{P^2\times D}$. Next, global representations $\mathbf{H} \newin \mathbb{R}^{L\times D}$ are obtained by spatially averaging and temporally concatenating $\mathbf{f}_i$. These $\mathbf{H}$ are then input to a frame-level enrichment (FLE, Sec.~\ref{sec:FLE}) block, which models higher-order temporal representations by aggregating the temporal context of actions across frames in a video. The resulting \textit{spatio-temporally enriched} features $\mathbf{E} \newin \mathbb{R}^{L\times D}$ of query and support videos are then input to the TRM, which models the temporal relationships between them. Moreover, a query-class similarity classifier (Sec.~\ref{sec:qcsim}) on the global representations $\mathbf{H}$ reinforces the network to learn class-discriminative features at different stages. Our framework is learned jointly using $\mathcal{L}_{TM}$ and $\mathcal{L}_{QC}$.\vspace{-0.3cm}}
    \label{fig:overall_arch}
\end{figure*}

\subsection{Baseline FS Action Recognition Framework\label{sec:trx}}
In this work, we adopt as baseline the recently introduced Temporal-relational CrossTransformer (TRX)~\cite{trx} method, which has shown to achieve state-of-the-art performance on multiple action recognition benchmarks. The TRX classifies a query video by matching it with the actions occurring at different speeds and instants in the support class videos using CrossTransformers~\cite{crosstransformers}. First, for each sub-sequence in the query video, a query-specific class prototype is computed via an aggregation of all possible sub-sequences in the support videos of an action class. The aggregation weights are based on the cross-attention values between the query sub-sequence and support class sub-sequences. Afterwards, the distances between the embeddings of the sub-sequences of a query video and their corresponding query-specific class prototypes are averaged to obtain the distance of the query to a class. 

The TRX method introduces hand-crafted representations to capture the higher-order temporal relationships, where sub-sequences are represented by tuples of different cardinalities based on the number of frames used for encoding a sub-sequence. For instance, with $\mathbf{e}_i\newin \mathbb{R}^D$ as the  $i^{th}$ frame representation, a sub-sequence between $t_i$ and $t_j$ can be represented as a pair ($\mathbf{e}_i$, $\mathbf{e}_j$) $\newin \mathbb{R}^{2D}$, a triplet ($\mathbf{e}_i$, $\mathbf{e}_k$, $\mathbf{e}_j$) $\newin \mathbb{R}^{3D}$, a quartet ($\mathbf{e}_i$, $\mathbf{e}_k$, $\mathbf{e}_l$, $\mathbf{e}_j$) $\newin \mathbb{R}^{4D}$ and so on, such that $1{\leq} i{<}k{<}l{<}j{\leq} L$. 
For a tuple  $t=(t_1,\cdots,t_\omega)$ of cardinality $\omega \newin \Omega$, let $\mathbf{q}_t \newin \mathbb{R}^{D^{'}}$ be a value embedding of query $Q_t=[\mathbf{e}_{t_1}; \cdots; \mathbf{e}_{t_\omega}] \newin \mathbb{R}^{\omega D}$ and $\mathbf{p}_t^c \newin \mathbb{R}^{D^{'}}$ be the query-cardinality-specific class prototype, obtained by the attention-based aggregation of value embeddings of support tuples $S^c_{kt} \newin \mathbb{R}^{\omega D}$. Then, the distance between a query video $Q$ and support set $\mathbf{S}^c$ over multiple cardinalities is given by,
\begin{equation}
\label{eqn:trx_method}
    \mathbf{T}(Q,\mathbf{S}^c) = \sum_{\omega \in \Omega} \frac{1}{|\Pi_\omega|} \sum_{t\in\Pi_\omega} \|\mathbf{q}_t - \mathbf{p}_t^c \|,
\end{equation}
where $\Pi_\omega{=}\{(t_1,\cdots,t_\omega) \in \mathbb{N}^\omega : 1 \leq t_1 < \cdots < t_\omega \leq L\} $ is the set of all possible tuples for cardinality $\omega$. The distance $\mathbf{T}(\cdot,\cdot)$ from a query video to its ground-truth class is minimized  by employing a standard cross-entropy loss during training. For further details, we refer to~\cite{trx}.

\vspace{1pt}\noindent\textbf{Limitations:} 
As discussed above, TRX performs temporal relationship modeling between the query and support action sub-sequences.
However, this modeling struggles in the case of spatial context variation (appearance change of relevant objects in query and support videos) and associated variation in temporal context (aggregation of spatial context across frames). Such variations are typically encountered in case of fine-grained action categories (see Fig.~\ref{fig:intro_fig}). 
Furthermore, TRX jointly employs multiple CrossTransformers, one for each different cardinality, to model higher-order temporal relationships based on different hand-crafted temporal representations of sub-sequences.
Consequently, this results in a less flexible model requiring dedicated branches for different cardinalities in addition to involving a manual model-search over different $\Omega$ combinations to find the optimal $\Omega^*$.
Next, we present our proposed approach that aims to collectively treat the aforementioned issues.

\section{Proposed STRM Approach}
\noindent\textbf{Motivation:} Here, we introduce our few-shot (FS) action recognition framework, STRM, which strives to enhance class-specific feature discriminability while simultaneously mitigating the flexibility issue. \\
\noindent\textit{Feature Discriminability:} Distinct from TRX that focuses \textit{solely} on temporal relationship modeling, our approach emphasizes the importance of aggregating spatial \textit{and} temporal context to effectively enrich the video sub-sequence representations before modeling the temporal relations. The local representation followed by learning rich spatial and temporal relationships enables enhanced feature discriminability, leading to an effective utilization of the limited samples available for FS action recognition. \\
\noindent\textit{Model Flexibility:} As discussed earlier, TRX employs hand-crafted higher-order temporal representations of different cardinalities, thereby requiring a search over multiple combinations. 
Instead, our approach learns to model higher-order relations at lower cardinalities with reduced inductive-bias, in turn improving the model flexibility. 

To collectively address both the above issues, we introduce an enrichment mechanism that targets enhanced feature discriminability of individual frames at a local patch-level (spatial) as well as the video itself at a global frame-level (temporal) while simultaneously learning higher-order temporal representations for improved flexibility.

\subsection{Overall Architecture\label{sec:overall_arch}}
Fig.~\ref{fig:overall_arch} illustrates our overall FS action recognition framework, STRM. The $L$ video frames are passed through an image-feature extractor, which outputs $D$-dimensional frame features with a spatial resolution $P{\times} P$. The frame features are then spatially flattened to obtain $ \mathbf{x}_i \newin \mathbb{R}^{P^2 \times D}$, $i \newin [1,L]$, which are then input to our novel spatio-temporal enrichment module comprising patch-level and frame-level enrichment sub-modules to obtain class-discriminative representations. 
The patch-level enrichment (PLE) sub-module enhances the patch features locally by attending to the spatial context in each frame and outputs spatially enriched features $\mathbf{f}_i\newin \mathbb{R}^{P^2 \times D}$ per frame. The $\mathbf{f}_i$'s are spatially averaged to obtain $D$-dimensional frame-level representations, which are then concatenated to form $\mathbf{H} \newin \mathbb{R}^{L\times D}$. Next, the frame-level enrichment (FLE) sub-module enhances the frame representations globally by encoding the temporal context from different frames in the video and outputs \textit{spatio-temporally enriched} frame-level representations $\mathbf{E} \newin \mathbb{R}^{L\times D}$. These representations $\mathbf{E}$ are input to a temporal relationship modeling (TRM) module, which classifies the query video by matching its sub-sequences with support actions. 
Additionally, classifying intermediate representations $\mathbf{H}$ by introducing a query-class similarity classifier reinforces the learning of corresponding class-level information at different stages and aids in further improving the overall feature discriminability. Our framework is learned jointly using standard cross-entropy loss terms $\mathcal{L}_{TM}$ and $\mathcal{L}_{QC}$ on the class predictions from the TRM module and query-class similarity classifier, respectively. Next, we present our proposed spatio-temporal enrichment module.

\subsection{Spatio-temporal Enrichment\label{sec:ste}}
The focus of our approach is the introduction of a spatio-temporal enrichment module that strives to enhance (i) local patch features spatially in an individual frame \textit{and} (ii) global frame features temporally across frames in a video. The effective utilization of both spatial as well as temporal contextual information within a video enables improved class-specific feature discriminability before modeling the temporal relationships between query and support videos.

\subsubsection{Enriching Local Patch Features \label{sec:PLE}}
\begin{figure}[t]
    \centering
    \includegraphics[width=1\columnwidth]{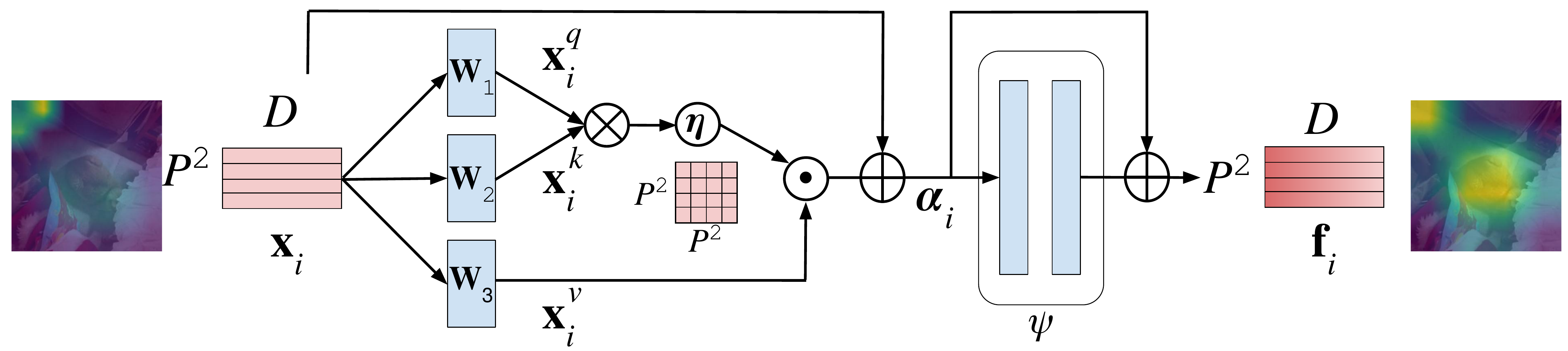}\vspace{-0.2cm}
    \caption{\textbf{Patch-level enrichment (PLE) sub-module.} Latent features $\mathbf{x}_i$ are projected by learnable weights $\mathbf{W}_1$, $\mathbf{W}_2$ and $\mathbf{W}_3$ to form query-key-value triplets ($\mathbf{x}_i^q$, $\mathbf{x}_i^k$, $\mathbf{x}_i^v$). 
    The value embeddings are re-weighted by the normalized pairwise scores between queries and keys, to obtain attended features ${\bm\alpha}_i$.
    A sub-network $\psi(\cdot)$ refines these ${\bm\alpha}_i$ to produce patch-level enriched features $\mathbf{f}_i$. Here, example attention maps before (on the left) and after (on the right) patch-level enrichment are shown. Best viewed zoomed in.\vspace{-0.3cm}}
    \label{fig:PLE}
\end{figure}

The patch features together in a frame encode its spatial information. Enhancing these features to encode the frame-level spatial context across all the patches in a frame is necessary to capture the appearance-based similarities as well as differences among the action classes. To this end, we introduce a patch-level enrichment (PLE) sub-module, which employs self-attention~\cite{vaswani2017attention} to let the patch features attend to themselves by aggregating the congruent patch contexts. The PLE sub-module is illustrated in Fig.~\ref{fig:PLE}. Let $\mathbf{x}_i \newin \mathbb{R}^{P^2 \times D}$ denote the latent features of $P^2$ patches in frame $q_i$ ($i\newin[1,L]$). Weights $\mathbf{W}_1$, $\mathbf{W}_2$, $\mathbf{W}_3 \newin \mathbb{R}^{D\times D}$ project these latent features to obtain query-key-value triplets, given by
\begin{equation}
  \mathbf{x}_i^q = \mathbf{x}_i\mathbf{W}_1, \quad \mathbf{x}_i^k =  \mathbf{x}_i\mathbf{W}_2, \quad \mathbf{x}_i^v = \mathbf{x}_i\mathbf{W}_3.
\end{equation}
While the \textit{value} embedding persists the current status of a patch $p\newin[1,P^2]$, the \textit{query} and \textit{key} vectors score the pairwise similarity between $P^2$ patches.
These \textit{value} embeddings are reweighted by corresponding normalized scores
to obtain `token-mixed' (attended) features ${\bm{\alpha}}_i$, given by
\begin{equation}
    {\bm\alpha}_i = \eta \left( \frac{\mathbf{x}_i^q \mathbf{x}_i^{k\top} }{\sqrt{D}} \right) \mathbf{x}_i^v  + \mathbf{x}_i,
\end{equation}
where $\eta$ denotes softmax function. 
A sub-network $\psi(\cdot)$ then point-wise refines these attended features ${\bm{\alpha}}_i \newin \mathbb{R}^{P^2 \times D}$ and outputs spatially enriched features $\mathbf{f}_i \newin \mathbb{R}^{P^2\times D}$, given by
\begin{equation}
    \mathbf{f}_i = \psi({\bm\alpha}_i) + {\bm\alpha}_i,
\end{equation}
leading to an improved aggregation of the appearance-based action context across patches in a frame (see Fig.~\ref{fig:method_viz} row 3).
%
%

\subsubsection{Enriching Global Frame Features \label{sec:FLE}}
\begin{figure}[t]
    \centering
    \includegraphics[width=1\columnwidth]{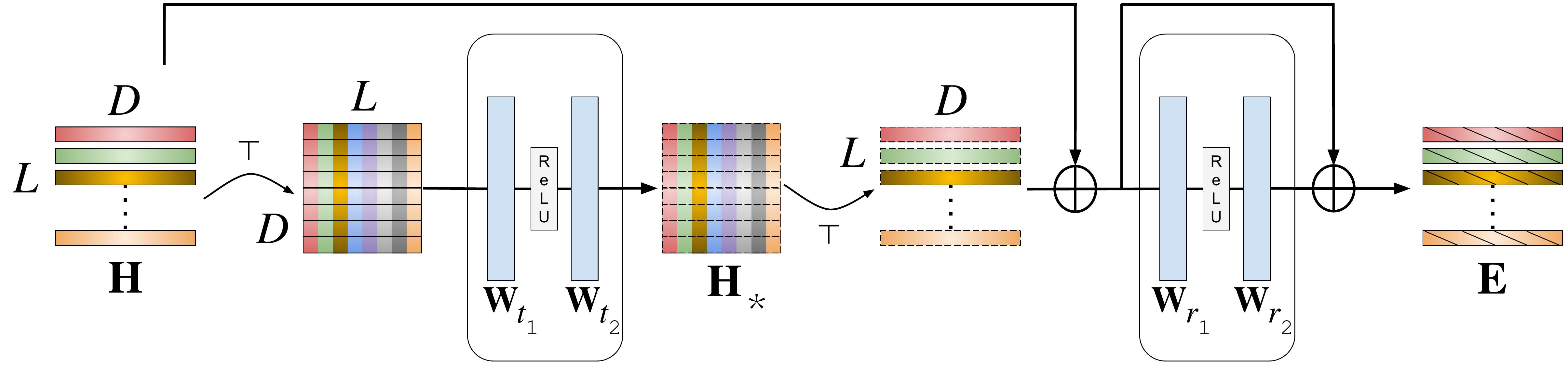}\vspace{-0.2cm}
    \caption{\textbf{Frame-level enrichment (FLE) sub-module.} The $L$ frame tokens $\mathbf{h}_i$ of global representation $\mathbf{H}$ are first mixed through learnable weights $\mathbf{W}_{t_1}$ and $\mathbf{W}_{t_2}$ that are shared across the feature dimensions $D$, to obtain intermediate representations $\mathbf{H}_*$. This is then followed by individual token refinement using weights $\mathbf{W}_{r_1}$ and $\mathbf{W}_{r_2}$ that are shared across the $L$ tokens, to obtain frame-level enriched features $\mathbf{E}$. Best viewed zoomed in. \vspace{-0.3cm}}
    \label{fig:FLE}
\end{figure}

\begin{figure*}[t]
    \centering
    \includegraphics[clip=true, trim=0em 0em 0em 0em, width=0.98\textwidth]{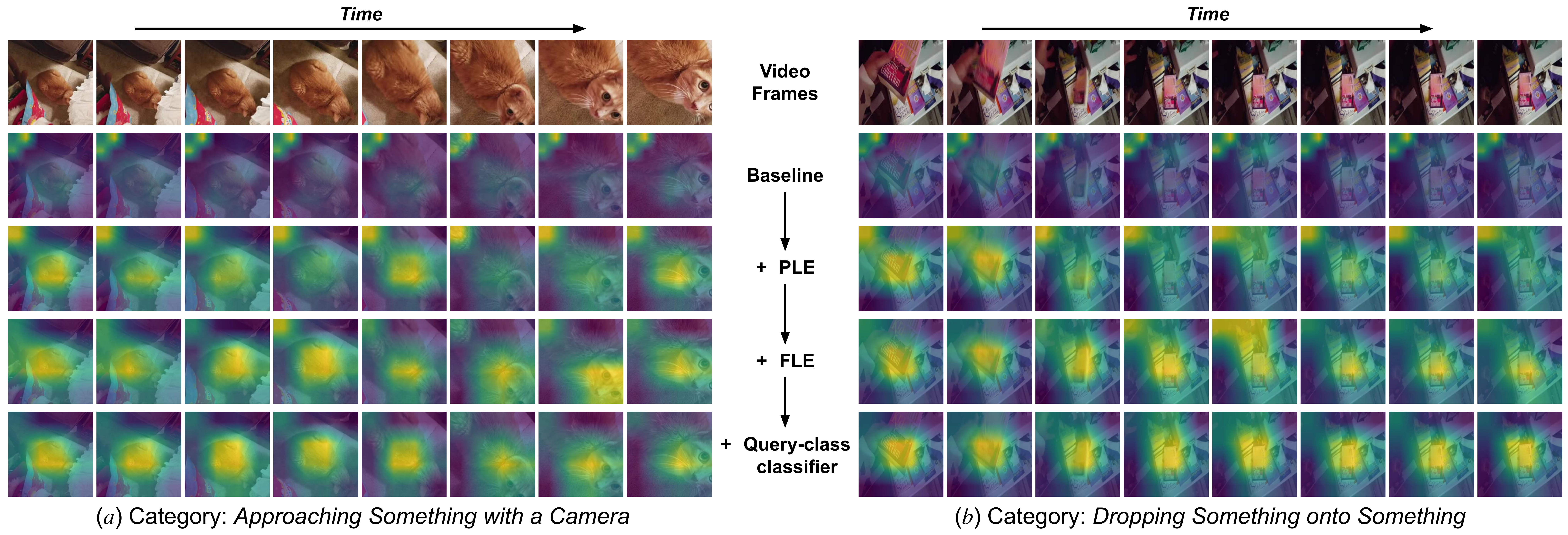}\vspace{-0.2cm}
    \caption{\textbf{The impact of progressively integrating our contributions} one at a time, from top (baseline) to bottom. The comparison is shown in terms of attention map visualizations measuring the activation magnitude of latent features for two examples from the SSv2~\cite{Goyal_2017_ICCV} test set. The baseline (second row) struggles to accurately capture the spatial as well as temporal contextual information. The integration of PLE sub-module (third row), which explicitly encodes spatial context, enables focus on relevant objects in a frame (second and third frame from the left in (a)). The integration of FLE sub-module (fourth row) further encodes the temporal context by consistently capturing the relevant object over time. For instance, while the context in fourth and sixth frame from the left in (a) are missed by PLE due to object motion, it is captured by the introduction of FLE. Lastly, integrating the query-class classifier further improves the attention on objects, leading to enhanced feature discriminability, \eg, seventh and eighth frame from the left in (b) has improved attention on the object (book).\vspace{-0.3cm}}
    \label{fig:method_viz}
\end{figure*}

The local patch-level enrichment (PLE) described above aims to aggregate the spatial contexts locally within each frame of an action video. This enables focusing on relevant objects in a frame. However, it does not explicitly encode the temporal context and therefore struggles when encountered with object motion over time (see Fig.~\ref{fig:method_viz}). 
Here, we proceed with the enrichment of temporal contexts globally across frames within a video by introducing a frame-level enrichment (FLE) sub-module comprising an MLP-mixer~\cite{mlp_mixer} layer. While self-attention is based on sample-dependent (input-specific) mixing guided by pairwise similarities between the tokens, the token-mixing in MLP-mixers assimilates the entire global receptive field through an input-independent and persistent relationship memory. Such a global assimilation of tokens enables the MLP-mixer to be better suited for enriching global frame representations. The FLE sub-module is shown in Fig.~\ref{fig:FLE}. 
For a frame $q_i$, let $\mathbf{h}_i \newin \mathbb{R}^D$ denote the global representation obtained by spatially averaging the PLE output $\mathbf{f}_i \newin \mathbb{R}^{P^2\times D}$. The concatenated global representation $\mathbf{H}=[\mathbf{h}_1;\cdots; \mathbf{h}_L]^\top \newin \mathbb{R}^{L\times D}$ for the entire video is then processed by the FLE sub-module. First, the frame tokens are mixed through a two-layer MLP $\mathbf{W}_t(\cdot)$ that is shared across channels (feature dimensions). This is followed by the token refinement of the intermediate features $\mathbf{H}_{*}$ by utilizing another two-layer MLP $\mathbf{W}_r(\cdot)$, which is shared across tokens. The two mixing operations in FLE are given by,
\begin{align}
   & \mathbf{H}_{*} =  \sigma( \mathbf{H}^\top\mathbf{W}_{t_1} ) \mathbf{W}_{t_2}  + \mathbf{H}^\top , \\
   & \mathbf{E}_{} = \sigma( \mathbf{H}_*^\top \mathbf{W}_{r_1} ) \mathbf{W}_{r_2}  + \mathbf{H}_*^\top ,
\end{align}
where $\mathbf{E} \newin \mathbb{R}^{L\times D}$ is the enriched feature, $\mathbf{W}_{t_1}, \mathbf{W}_{t_2} \in \mathbb{R}^{L\times L}$ and $\mathbf{W}_{r_1}, \mathbf{W}_{r_2} \in \mathbb{R}^{D\times D}$ are the learnable weights for token- and channel-mixing, respectively. Here, $\sigma$ denotes the ReLU non-linearity. In particular, the token-mixing operation ensures that the frame representations interact together and imbibe the higher-order temporal relationships through the learnable weights $\mathbf{W}_{t_1}$ and $\mathbf{W}_{t_2}$. As a result, the FLE sub-module enhances the frame representations $\mathbf{h}_i$ temporally, with a global receptive field encompassing all the frames and produces temporally-enriched representations $\mathbf{e}_i$ for $i\in[1,L]$.

The enriched frame-level global representations $\mathbf{e}_i$  ($i\newin[1,L]$) for the query and support videos are then input to the temporal relationship modeling (TRM) module, which models the temporal relationships between query and support actions. Within our framework, the TRM is a TRX (Eq.~\ref{eqn:trx_method}) built on a single cardinality $\Omega \neweq \{2\}$, since our spatio-temporal enrichment module \textit{learns} to model higher-order temporal representations without requiring multiple hand-crafted cardinality representations. Given the ground-truth labels $\mathbf{y}\in\mathbb{R}^C$, our framework is then learned end-to-end using the standard cross-entropy (CE) loss on the class probabilities $\hat{\mathbf{y}}_{TM} \in \mathbb{R}^C$ predicted by the TRM, given by 
\begin{equation}
\label{eqn:trxloss}
\mathcal{L}_{TM} = \mathbb{E} [\text{CE}(\hat{\mathbf{y}}_{TM}, \mathbf{y})].
\end{equation}
In summary, our spatio-temporal enrichment module leverages the advantages of local and global, sample-dependent and sample-agnostic enrichment mechanism to  improve the aggregation of spatial as well as temporal contexts of actions. As a result, class-specific discriminative features are obtained along with the assimilation of higher-order temporal relationships in lower cardinality representations.

\subsection{Query-class Similarity \label{sec:qcsim}}
As discussed above, the proposed framework comprising the feature extractor, spatio-temporal enrichment and temporal relationship modeling modules, is learned end-to-end with a CE loss on the output probabilities $\hat{\mathbf{y}}_{TM}$. However, learning to classify query video representations from intermediate layer outputs reinforces the model to look for class-specific features at different stages in the pipeline. Consequently, such a multi-stage classification improves the feature discriminability, leading to better matching between query and support videos. To this end, we introduce a query-class similarity classifier on the patch-level enriched representations $\mathbf{h}_i$, $i\newin[1,L]$. 
First, we obtain latent tuple representations $\mathbf{l}_t \neweq [\mathbf{h}_{t_1}; \cdots; \mathbf{h}_{t_\omega}] \in \mathbb{R}^{\omega D}$ for tuples $t \neweq (t_1,\cdots,t_\omega) \in \Pi_\omega$ in a video. 
They are then projected by $\mathbf{W}_{cls} \newin \mathbb{R}^{\omega D \times D^{''}}$ to obtain $\mathbf{z}_t \neweq \sigma(\mathbf{W}_{cls}^\top \mathbf{l}_t)$, where $\sigma$ is the ReLU non-linearity. 
Then, for each $\mathbf{z}_t^Q$ in a query video $Q$, its highest similarity among all tuples in the $K$ support videos for an action class $c$ is computed. These scores for all the tuples in $Q$ are aggregated to obtain the query-class similarity $M(Q,c)$ between the query and action $c$. 
With $\mathbf{z}^{c}_j$ representing a tuple $j \in [1, K\cdot|\Pi_\omega|]$ from the $K$ support videos for an action $c$, the query-class similarity is given by
\begin{equation}
    M(Q, c) = \sum_{\omega \in \Omega} \frac{1}{|\Pi_\omega|}\sum_{t\in \Pi_\omega} \max_{j} \phi(\mathbf{z}^Q_t, \mathbf{z}^{c}_j),
\end{equation}
where $\phi(\cdot,\cdot)$ is a similarity function. Then, the $C$ similarity scores are passed through \textit{softmax} to obtain class probabilities $\hat{\mathbf{y}}_{QC} \newin \mathbb{R}^C$ and trained with a CE loss given by
\begin{equation}
     \mathcal{L}_{QC} = \mathbb{E} [\text{CE}(\hat{\mathbf{y}}_{QC}, \mathbf{y})]. 
\end{equation}
With $\lambda$ as a hyper-weight, our STRM is trained using the joint formulation given by
\begin{equation}
    \mathcal{L} = \mathcal{L}_{TM} + \lambda\mathcal{L}_{QC}.
\end{equation}
Consequently, our proposed STRM, comprising a  spatio-temporal enrichment module and an intermediate query-class similarity classifier, enhances feature discriminability (see Fig.~\ref{fig:method_viz}) and leads to improved matching between queries and their support action classes.

\section{Experiments}
\noindent\textbf{Datasets:} Our approach is evaluated on four popular benchmarks: Something-Something V2 (SSv2) \cite{Goyal_2017_ICCV}, Kinetics \cite{carreira2017quo}, HMDB51~\cite{hmdb} and UCF101 \cite{ucf101}. The SSv2 is crowd-sourced, challenging and has actions requiring temporal reasoning. For SSv2, we use the split with $64$/$12$/$24$ action classes in training/validation/testing, given by~\cite{otam}. A similar split with $64$/$12$/$24$ action classes, as in~\cite{cmn,otam} is used for Kinetics. Furthermore, we evaluate on HMDB51 and UCF101 using the splits from~\cite{arn}. The standard $5$-way $5$-shot evaluation is employed on all datasets and the average accuracy over $10{,}000$ random test tasks is reported. \\
\noindent\textbf{Implementation Details:} As in \cite{trx,otam}, a ResNet-$50$ \cite{resnet}, pretrained on ImageNet~\cite{imagenet}, is used as the feature extractor for $L\neweq 8$ uniformly sampled frames of a video. With $D\neweq 2{,}048$, an adaptive maxpooling reduces the spatial resolution to $P\neweq 4$. All the learnable weights matrices in PLE and FLE are implemented as fully-connected (FC) layers. The sub-network $\psi(\cdot)$ in PLE is a $3$-layer FC network with latent sizes set to $1{,}024$. We set $D^{''}\neweq 1{,}024$ for $\mathbf{W}_{cls}$. For the TRM, we employ $\Omega\neweq \{2\}$ in Eq.~\ref{eqn:trx_method} and set $D^{'}\neweq 1{,}152$, as in~\cite{trx}. The hyper-weight $\lambda$ is set to $0.1$. While $75{,}000$ randomly sampled training episodes are used for SSv2 dataset with a learning rate of $10^{-3}$, the smaller datasets are trained with a $10^{-4}$ learning rate. Our STRM framework is trained end-to-end using an SGD optimizer.

%
%
\begin{table}[t]
\centering
\caption{\label{tab:sota_comparison} \textbf{State-of-the-art comparison on four FS action recognition datasets}, in terms of classification accuracy. Our \texttt{STRM} outperforms existing FS action recognition methods on all four datasets. Importantly, for ResNet-50 backbone, \texttt{STRM} achieves an absolute gain of 3.5\% over TRX~\cite{trx} on the challenging SSv2 that comprises actions requiring temporal relationship reasoning.\vspace{-0.3cm}}
\setlength{\tabcolsep}{8pt}
\adjustbox{width=\linewidth}{
 
 \begin{tabular}{lccccc}
 \toprule[0.1em]
 \textbf{Method} & \textbf{Backbone} & \textbf{Kinetics} & \textbf{SSv2} & \textbf{HMDB} & \textbf{UCF}\\
 \toprule[0.1em]
  \texttt{CMN-J}~\cite{cmn} & ResNet-50 & 78.9 & - & - & - \\
  \texttt{TARN}~\cite{tarn} & ResNet-50 & 78.5 & - & - & - \\
  \texttt{ARN}~\cite{arn} & ResNet-50 & 82.4 & - & 60.6 & 83.1 \\
  \texttt{OTAM}~\cite{otam} & ResNet-50 & 85.8 & 52.3 & - & - \\
  \texttt{HF-AR}~\cite{kumar2021shot} & ResNet-50 & - & 55.1 & 62.2 & 86.4 \\
  \texttt{TRX}~\cite{trx} & ResNet-50 & 85.9 & 64.6 & 75.6 & 96.1 \\
  \textbf{\texttt{Ours:STRM}} & ResNet-50 & \textbf{86.7} & \textbf{68.1} & \textbf{77.3} & \textbf{96.9} \\
  \midrule
  \texttt{TRX}~\cite{trx} & ViT & 90.6 &  67.3 & 79.7 & 97.1 \\
  \textbf{\texttt{Ours:STRM}} & ViT & \textbf{91.2} & \textbf{70.2} & \textbf{81.3} & \textbf{98.1}\\
  \bottomrule[0.1em]
 \end{tabular}
 
}
\vspace{-0.2cm}
\end{table}

\subsection{State-of-the-art Comparison}
Tab.~\ref{tab:sota_comparison} shows the state-of-the-art comparison on four benchmarks for the standard $5$-way $5$-shot action recognition task. For fairness, only the approaches employing a 2D backbone for extracting per-frame features are compared in Tab.~\ref{tab:sota_comparison}.
On Kinetics, the recent works of \texttt{OTAM}~\cite{otam} and \texttt{TRX}~\cite{trx} achieve comparable classification accuracies of $85.8$ and $85.9\%$. Our \texttt{STRM} performs favorably against existing methods by achieving an improved performance of $86.5\%$. On the more challenging SSv2 dataset comprising actions requiring temporal relational reasoning, \texttt{OTAM} and \texttt{HF-AR}~\cite{kumar2021shot} achieve $52.3\%$ and $55.1\%$, while \texttt{TRX} obtains an accuracy of $64.6\%$, due to its temporal relationship modeling. Compared to the best existing approach of \texttt{TRX}, our \texttt{STRM} achieves a significant absolute gain of $3.5\%$ on SSv2. Similarly, our \texttt{STRM} achieves improved performance on HMDB51 and UCF101, setting a new state-of-the-art on \textit{all} four benchmarks.
To further evaluate our contributions, we replace the ResNet-50  with ViT~\cite{dosovitskiy2020image} as the backbone. Even with this stronger backbone, our \texttt{STRM} outperforms \texttt{TRX} on all datasets. In addition, our \texttt{STRM} achieves gains of 1.5\% and 1.9\% over \texttt{TRX} on SSv2, when employing 3D ResNet-50 and MViT~\cite{fan2021multiscale} backbones. Note that the 3D ResNet-50 and MViT are pretrained on Kinetics400~\cite{carreira2017quo} and are \textit{not always} compatible with few-shot action datasets due to possible overlap of pretraining classes with novel classes. 
The consistent improvement of our \texttt{STRM} emphasizes the efficacy of enhancing spatio-temporal features, by integrating local (sample-dependent) patch-level and global (sample-agnostic) frame-level enrichment along with a query-class similarity classifier, for FS action recognition.
\begin{figure}[t]
    \centering
    \includegraphics[clip=true, trim=2em 2em 6em 3em, width=0.7\columnwidth]{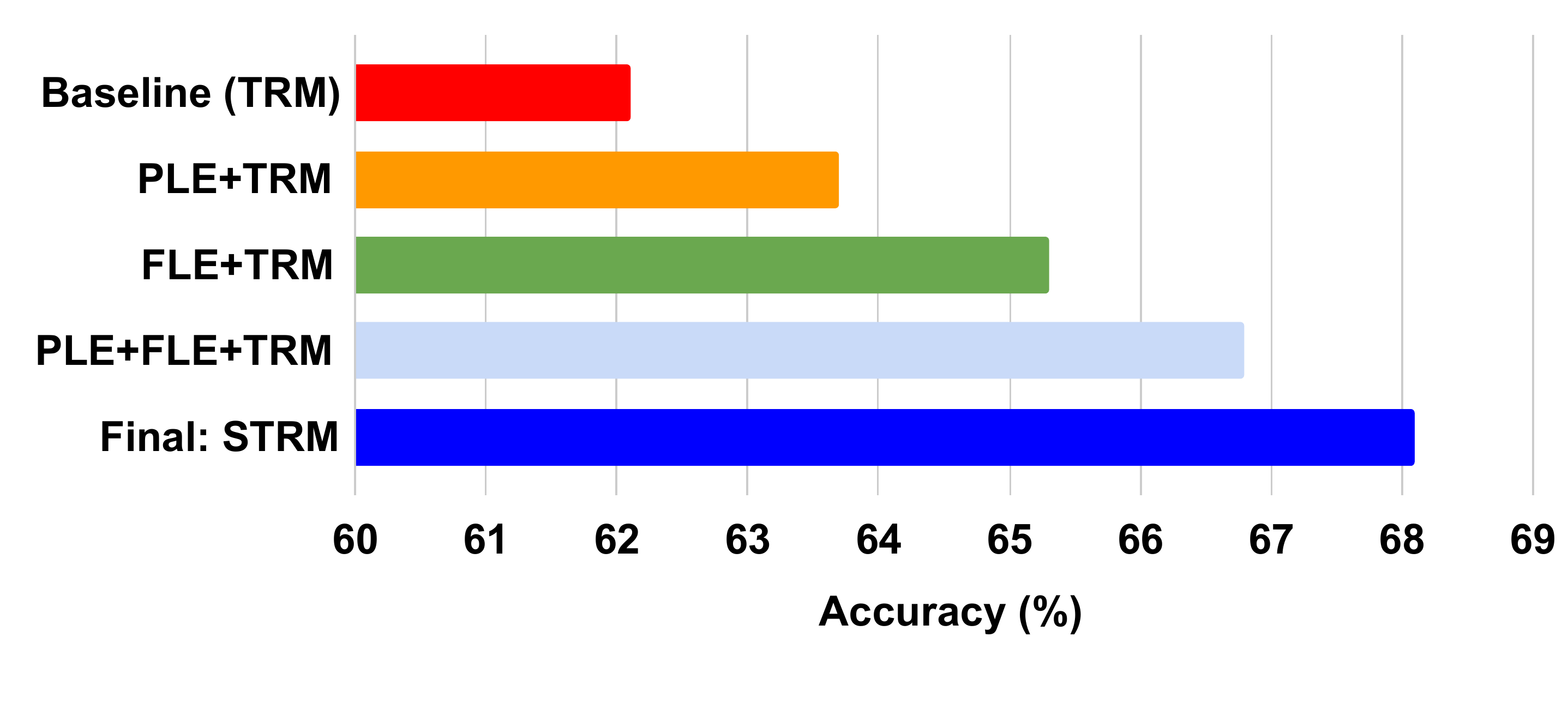} \hspace{0.2cm}
    \includegraphics[clip=true, trim=2em 2em 1em 2em, width=0.25\columnwidth]{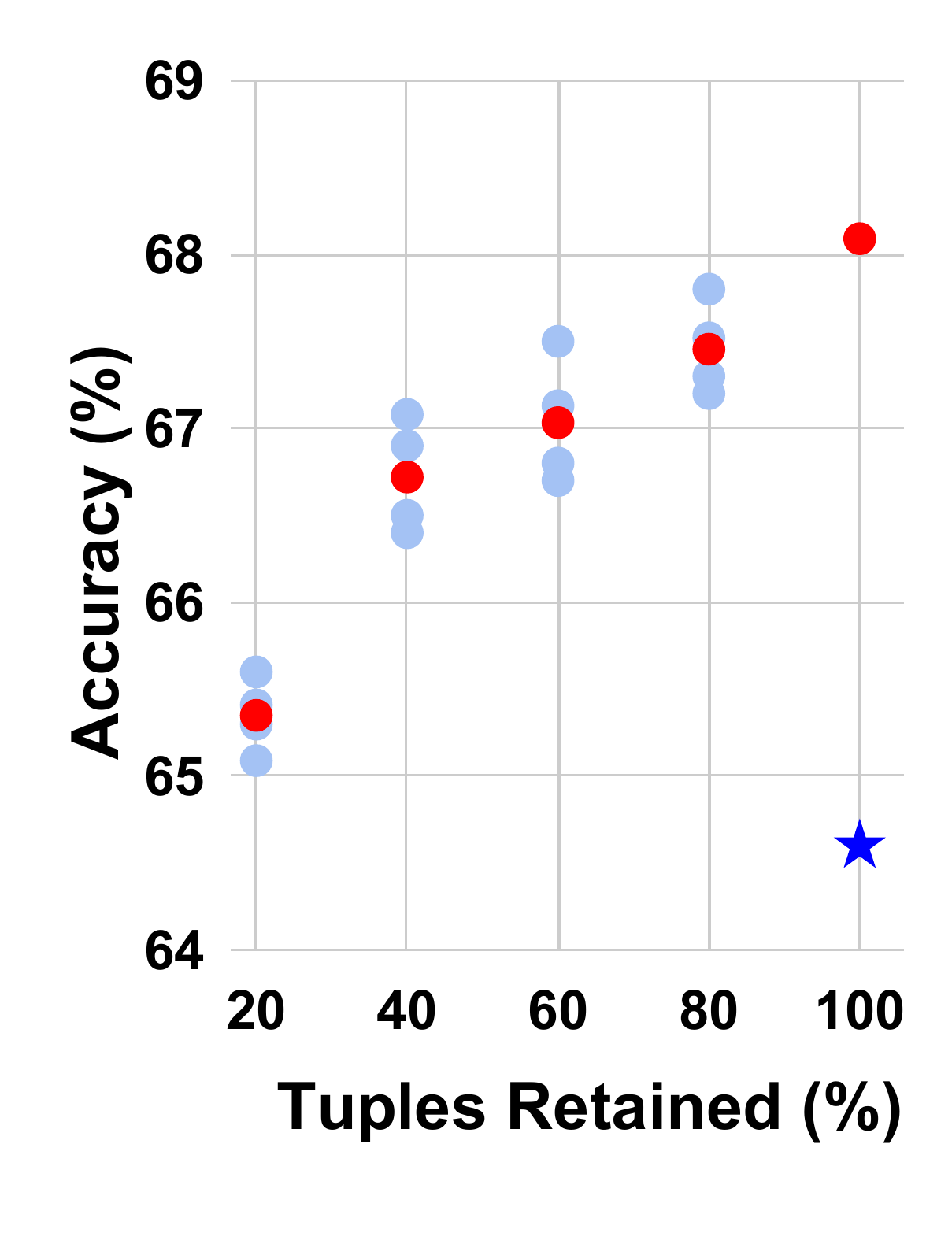} \vspace{-0.7cm}
    \caption{\textbf{(Left) Impact of integrating our contributions in the baseline} on SSv2.
    Individually integrating our PLE (\textcolor{orange}{orange bar}) and FLE (\textcolor{mygreen}{green bar}) into the baseline TRM results in improved performance. The joint integration (\textcolor{ProcessBlue}{light blue bar}) of PLE and FLE in the baseline enriches spatio-temporal features, leading to superior performance. Lastly, integrating our query-class classifier further enhances the feature discriminability. Our final \texttt{STRM} (\textcolor{blue}{blue bar}) obtains an absolute gain of 6.0\% over baseline.\\
    \textbf{(Right) Impact of varying \#tuples in our STRM}. Multiple trials and mean performance of \texttt{STRM} are denoted by $\bluecirc$ and $\redcirc$, respectively. Since the feature discriminability is enhanced due to spatio-temporal enrichment, even with only $20\%$ tuples retained,  \texttt{STRM} ($\Omega\neweq\{2\}$) performs favorably against \texttt{TRX} (denoted by \textcolor{blue}{$\bigstar$}) using $\Omega\neweq\{2,3\}$ and retaining all tuples. Best viewed zoomed in.\vspace{-0.35cm}}
    \label{fig:ablation}
\end{figure}

\subsection{Ablation Study}

\noindent\textbf{Impact of the proposed contributions:} Here, we systematically analyse the impact of our spatio-temporal enrichment module along with the query-class classifier. Note that our spatio-temporal enrichment module comprises PLE and FLE sub-modules. Fig.~\ref{fig:ablation} (left) shows a performance comparison on SSv2, when integrating our two contributions (spatio-temporal enrichment module and the query-class classifier) in the baseline TRM. Note that the baseline TRM is a TRX~\cite{trx} with cardinality $\Omega\neweq\{2\}$. The baseline TRM achieves an FS action classification accuracy of $62.1\%$ (\textcolor{red}{red bar}). Integrating our PLE in the baseline, for enriching the spatial context in the local patch-level features before temporal modeling, achieves an improved accuracy of $63.7\%$ (\textcolor{orange}{orange bar}). Similarly, enriching the temporal context alone in the global frame-level features through the integration of FLE  (\textcolor{mygreen}{green bar}) in TRM achieves a gain of $3.2\%$. 
Moreover, the joint integration of PLE and FLE (\textcolor{ProcessBlue}{light blue bar}) in the TRM further enhances the spatio-temporal contexts in the features, leading to an improved accuracy of $66.8\%$. 
Lastly, integrating the query-class classifier in our approach reinforces the learning of class-separable features at different stages and further enhances feature discriminability, thus, achieving a superior performance of $68.1\%$. The final \texttt{STRM} framework (\textcolor{blue}{blue bar}) achieves an absolute gain of 6.0\% over the baseline (\textcolor{red}{red bar}).\\
\noindent\textbf{Impact of varying cardinalities:} Tab.~\ref{tab:vary_omega} shows the impact of varying the cardinalities considered for modeling temporal relationships in our \texttt{STRM}. The comparison is shown for Kinetics and SSv2. The number of tuples present in corresponding cardinality combinations is also shown. We observe that our \texttt{STRM} achieves optimal performance even at lower cardinalities. In particular, our \texttt{STRM} achieves the best performance on both datasets with $\Omega\neweq\{2\}$. In contrast, \texttt{TRX} employing hand-crafted higher-order temporal representations requires $\Omega\neweq\{2,3\}$ to achieve its optimal performance of $64.6\%$ on SSv2. Moreover, it is worth mentioning that our \texttt{STRM} is comparable to \texttt{TRX} in terms of compute, requiring only $\!\sim\!4\%$ additional FLOPs. The superior performance of our approach over \texttt{TRX} at lower cardinality is due to the enhanced feature discriminability achieved through the spatio-temporal feature enrichment and the learning of higher-order temporal representations caused by token-mixing in our FLE sub-module.
\begin{table}[t]
\centering
\caption{\label{tab:vary_omega}\textbf{Impact of varying the cardinalities for temporal relationships in our \texttt{STRM}} on Kinetics and SSv2. Here, we also show the number of tuples available in the corresponding cardinality combinations. Our \texttt{STRM} achieves best performance at a lower cardinality of $\Omega\neweq\{2\}$, thereby mitigating the need of multiple TRM branches for different cardinalities.\vspace{-0.3cm}}
\adjustbox{width=\linewidth}{
\begin{tabular}{lcccccccc}
\toprule[0.1em]
\textbf{Cardinalities} ({$\bm\Omega$}) & $\{1\}$ & $\{2\}$ & $\{3\}$ & $\{4\}$ & $ \{2,3\}$ & $ \{2,4\}$ & $ \{3,4\}$ & $\{2,3,4\}$ \\ 
$\#$\textbf{Tuples} & - & 28 & 56 & 70 & 84 & 98 & 126 & 154 \\
\midrule
\textbf{Kinetics} & 86.2 & \textbf{86.5} & 86.0 & 85.3 & 85.9 & 86.1 & 85.7 & 86.1 \\
\textbf{SSv2} & 67.2 & \textbf{68.1} & 66.9 & 66.4 & 67.1 & 67.3 & 67.3 & \textbf{68.1} \\
\bottomrule[0.1em]
    \end{tabular}

}
\vspace{-0.2cm}
\end{table}

\begin{figure}[t]
\centering
    \includegraphics[clip=true, trim=2em 4.5em 2em 2em, width=0.85\columnwidth]{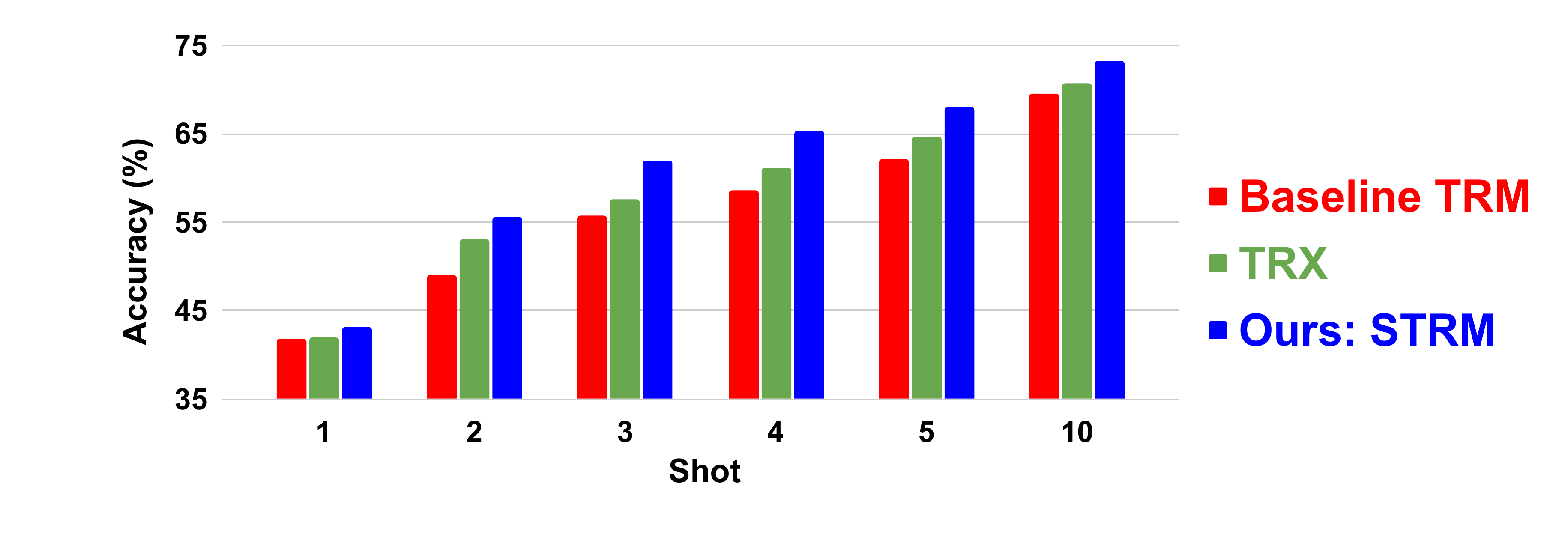}\vspace{-0.2cm}
    \caption{\label{fig:lowshot}\textbf{Performance comparison when varying the number of support samples} in the SSv2 dataset. We show the comparison of our \texttt{STRM} with both TRM and TRX. \texttt{STRM} achieves superior performance compared to both TRM and TRX on all settings, including the challenging one-shot case. Furthermore, \texttt{STRM} effectively leverages larger support set in the $10$-shot settings.\vspace{-0.3cm} }
\end{figure}

\noindent\textbf{Impact of varying tuples:} Fig.~\ref{fig:ablation} (right) shows the performance of our \texttt{STRM} approach on SSv2 when retaining different number of tuples for matching between query and support videos. We observe a marginal drop when the retained tuples are decreased. 
Moreover, even when retaining only $20\%$ of the tuples at a lower cardinality ($\Omega\neweq\{2\}$), our \texttt{STRM} achieves an accuracy of $65.4\%$ and performs favorably against $64.6\%$ of TRX, which relies on all the tuples from multiple cardinalities ($\Omega\neweq\{2,3\}$). 
This shows that our spatio-temporal enrichment module along with the query-class classifier enhances the feature discriminability while learning higher-order temporal representations in lower cardinalities itself. As a result, our \texttt{STRM} provides improved model flexibility, without requiring dedicated TRM branches for different cardinalities. \\
\noindent\textbf{Comparison with different number of support samples:} Fig.~\ref{fig:lowshot} compares \texttt{STRM} with the baseline and the TRX, when varying the number of support samples on SSv2. Here, we show $K$-shot ($K\!\leq\!5$ and $10$) classification. Our \texttt{STRM} achieves consistent improvement in performance, compared to both TRM and TRX on all $K$-shot settings. Specifically, our \texttt{STRM} excels in the extreme one-shot case as well as the $10$-shot setting, where it effectively leverages larger support sets.
Additional results are provided in the appendix.

\section{Relation to Prior Art\label{sec:related_work}}
Several works have investigated the few-shot (FS) problem for image classification~\cite{finn2017model,crosstransformers,bateni2020improved}, object detection~\cite{karlinsky2019repmet,wang2020frustratingly}, and segmentation~\cite{liu2020crnet}. 
While earlier approaches were either adaptation-based~\cite{nichol2018first}, generative~\cite{zhang2018metagan}, or metric-based~\cite{snell2017prototypical,vinyals2016matching}, recent works~\cite{requeima2019fast,crosstransformers} employ a combination of these.
In the context of FS action recognition, \cite{Zhu_2018_ECCV,cmn} employ memory networks for key-frame representations, whereas \cite{tarn} aligns variable length query and support videos. Differently, \cite{otam} utilizes monotonic temporal ordering for enforcing temporal consistency between video pairs. 
The recent work of TRX \cite{trx} focuses on modeling the temporal relationships by utilizing fixed higher-order temporal representations. Distinct from TRX, our STRM introduces a spatio-temporal enrichment module to produce spatio-temporally enriched features. The spatio-temporal enrichment module enriches features at local patch-level by employing a self-attention layer \cite{vaswani2017attention, Felix2019context, Ramachandran2019context} as well as global frame-level by utilizing an MLP-mixer layer \cite{mlp_mixer,tay2021synthesizer,touvron2021resmlp}.  Our spatio-temporal enrichment also enables learning higher-order temporal representations at lower cardinalities. 
The proposed spatio-temporal enrichment module performs local patch-level enrichment using a self-attention layer as well as global frame-level enrichment by integrating a MLP-mixer, in a FS action recognition framework.  Furthermore, we introduce a query-class classifier for learning to classify feature representations from intermediate layers.

\section{Discussion}
We proposed a FS action recognition framework, STRM, comprising spatio-temporal enrichment and temporal relationship modeling (TRM) modules along with a query-class similarity classifier. 
Our STRM leverages the advantages of combining local and global, sample-dependent and sample-agnostic enrichment mechanism for enhancing the spatio-temporal features, in addition to reinforcing class-separability of features at different stages. Consequently, this enhances the spatio-temporal feature discriminability and enables the learning of higher-order temporal relations even in lower cardinality representations. Our extensive ablations reveal the benefits of the proposed contributions, leading to state-of-the-art results on all benchmarks. 
A likely future direction, beyond the scope of the current work, is to broaden the few-shot action recognition capability to generalize across varying domains.

\section*{Acknowledgements}
This work was partially supported by VR starting grant (2016-05543), in addition to the compute support provided at the Swedish National Infrastructure for Computing (SNIC), partially funded by the Swedish Research Council through grant agreement 2018-05973.

\section*{Appendix}
In this appendix, we present additional quantitative (Sec.~\ref{sec:quant}) and qualitative (Sec.~\ref{sec:qual}) results of our proposed few-shot (FS) action recognition framework, STRM. 

\appendix

\section{Additional Quantitative Results\label{sec:quant}}

\noindent\textbf{Impact of joint spatio-temporal enrichment:} Tab.~\ref{tab:joint_sa} shows the impact of replacing our patch-level enrichment (PLE) and frame-level enrichment (FLE) sub-modules in the proposed \texttt{STRM} framework with a joint spatio-temporal (\texttt{Jnt-ST}) enrichment sub-module on the SSv2~\cite{Goyal_2017_ICCV} dataset. The performance of \texttt{Baseline TRM} is also shown for comparison. Jointly enriching all the spatio-temporal patches across the frames, as in \texttt{Jnt-ST}, does improve the performance over the baseline but with a $50\%$ increase in FLOPs due to computing attention over all the spatio-temporal patches in a video. Although using two layers of \texttt{Jnt-SA} gains over the single layer variant, it requires twice the number of FLOPs than \texttt{Baseline TRM}. Our proposed approach of enriching patches locally with in a frame and then enriching the frames globally in a video requires only $\sim 4\%$ additional FLOPs over the baseline and obtains superior performance. This shows the importance of proposed enrichment mechanism in our \texttt{STRM} framework.

\noindent\textbf{Impact of varying the enrichment mechanism:} We present the impact of varying the enrichment mechanisms in our PLE and FLE sub-modules in Tab.~\ref{tab:varying_enrichment} on the SSv2 dataset. It is worth mentioning that irrespective of the enrichment mechanism employed, integrating PLE and FLE sub-modules enhances the feature discriminability, leading to improved performance over \texttt{Baseline TRM}. However, we observe that employing an MLP-mixer~\cite{mlp_mixer} for enriching patches locally with in a frame (PLE) or employing self-attention~\cite{vaswani2017attention} for enriching frames globally across frames in a video (FLE) results in sub-optimal performance. This is because self-attention enriches the tokens locally in a pairwise and sample-dependent manner and is likely to be less suited for enriching the frames at a global level. Similarly, the MLP-mixer is sample-agnostic and enriches the tokens globally through a persistent relationship memory while being less suitable for enriching the patches at a local level.

\begin{table}[t]
    \centering
    \caption{\textbf{Impact of replacing our PLE and FLE sub-modules with joint spatio-temporal self-attention sub-module on SSv2.} Enriching all the spatio-temporal patches jointly across frames, denoted by \texttt{Jnt-ST} (number of layers shown in parenthesis), improves over \texttt{Baseline TRM}. 
    However, enriching patches spatially at a local level followed by enriching frames temporally at a global level in a hierarchical fashion, as in our \texttt{STRM}, obtains superior performance. 
    \vspace{-0.3cm}}
    \setlength{\tabcolsep}{6pt}
    \adjustbox{width=1\columnwidth}{
    \begin{tabular}{cccc}
    \toprule[0.1em]
    \texttt{Baseline TRM} & \texttt{Jnt-ST} (1 \textit{l}) & \texttt{Jnt-ST} (2 \textit{l}) & \textbf{\texttt{Ours:STRM}} \\
    \toprule[0.1em]
    62.1 & 64.7 & 65.8 & \textbf{68.1} \\
    \bottomrule[0.1em]
    \end{tabular}
    }
    \label{tab:joint_sa}
\end{table}

\begin{table}[t]
    \centering
    \caption{\textbf{Impact of varying the enrichment mechanism in PLE and FLE sub-modules of our STRM on SSv2.} The enrichment mechanism at patch-level and frame-level are varied between self-attention and MLP-mixer based implementations. The performance of \texttt{Baseline TRM} without any PLE and FLE is also shown for comparison. Irrespective of the enrichment mechanism employed, integrating PLE and FLE sub-modules improves over the baseline performance. Employing either MLP-mixer for local patch-level enrichment or self-attention for global frame-level enrichment yields sub-optimal performance. The best performance is obtained by our \texttt{STRM} when self-attention based PLE and MLP-mixer based FLE are integrated in the framework. \vspace{-0.3cm}}
    \setlength{\tabcolsep}{10pt}
    \adjustbox{width=1\columnwidth}{
    \begin{tabular}{cccc}
    \toprule[0.1em]
    & PLE & FLE & Accuracy\\
    \toprule[0.1em]
    \multirow{1}{*}{\texttt{Baseline TRM}} & - & - & 62.1 \\
    \midrule
    \multirow{4}{*}{\textbf{\texttt{Ours:STRM}}} & Self-attention & Self-attention & 64.2 \\

    & MLP-Mixer & Self-attention & 64.1 \\

    & MLP-Mixer & MLP-Mixer & 65.0 \\
    
     & Self-attention & MLP-Mixer & \textbf{68.1} \\
    \bottomrule[0.1em]
    \end{tabular}
    }
    \label{tab:varying_enrichment}
\end{table}

Thereby, employing self-attention for local patch-level enrichment and simultaneously an MLP-mixer for global frame-level enrichment achieves the best performance and achieves an absolute gain of $6.0\%$ over baseline. These results emphasize the efficacy of enhancing spatio-temporal features by integrating local (sample-dependent) patch-level and global (sample-agnostic) frame-level enrichment along with a query-class similarity classifier in our \texttt{STRM} for the task of FS action recognition.

\begin{figure}[t]
    \centering
    \includegraphics[clip=true, trim=0em 3em 0em 0em, width=0.8\columnwidth]{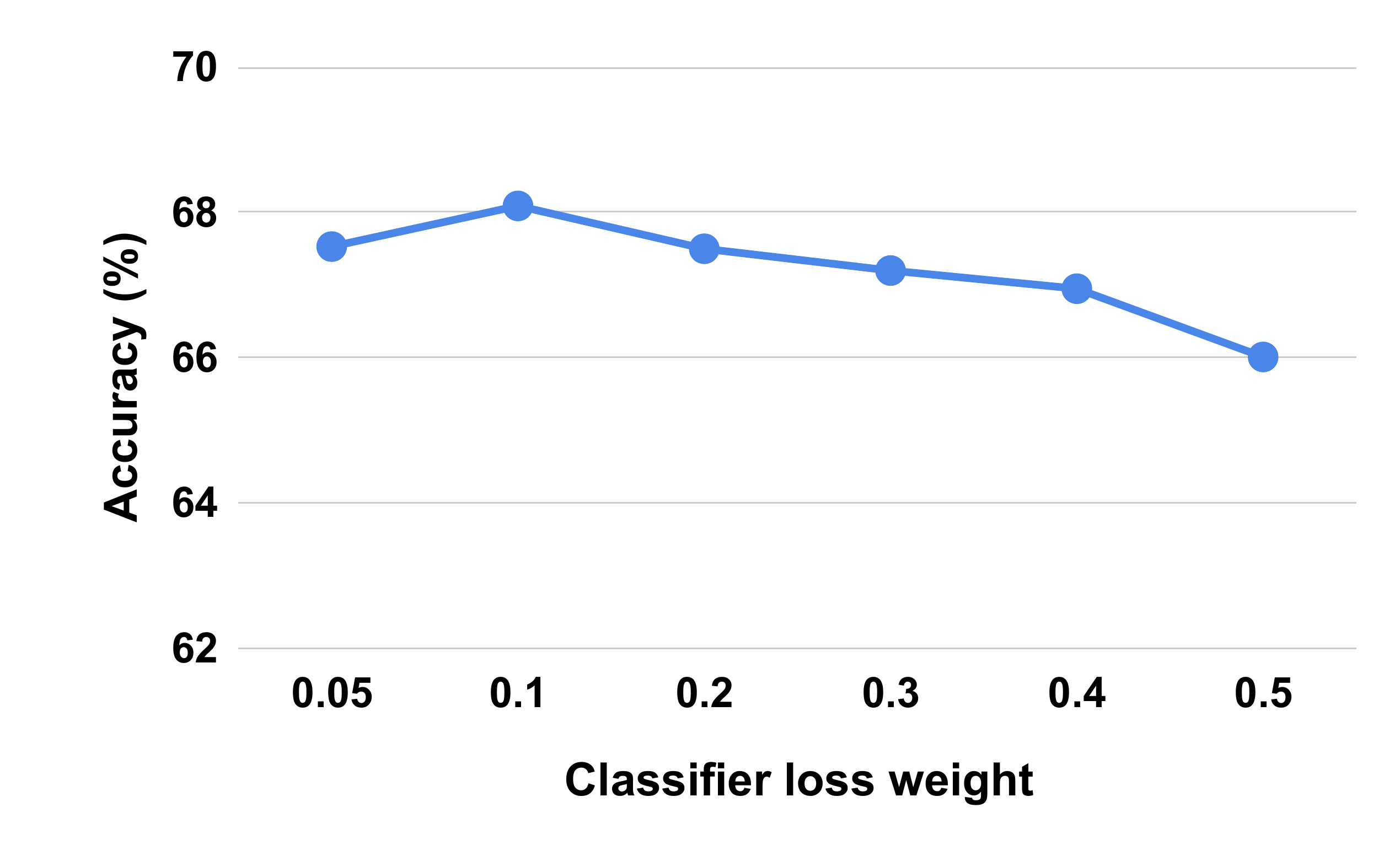}\vspace{-0.2cm}
    \caption{\textbf{Impact of varying $\lambda$ on SSv2.} A low weight for the query-class similarity classification loss yields the best performance for our \texttt{STRM} framework. Training with a large weight ($>0.4$) for this auxiliary classification loss decreases the importance of modeling temporal relationships in the TRM module and negatively affects the performance.}
    \label{fig:lambda_vary}
\end{figure}

\noindent\textbf{Impact of varying $\lambda$:} Fig.~\ref{fig:lambda_vary} shows the FS action recognition performance comparison for different values of $\lambda$, which is the weight factor for the query-class similarity classification loss in the proposed \texttt{STRM} framework.  Setting $\lambda$ high ($>0.4$) is likely to decrease the importance of the modeling temporal relationships between query and support actions in the TRM module during training and consequently leads to a drop in performance. Furthermore, we observe that employing this intermediate layer classification loss with a low weight (around $0.1$) improves the performance and achieves the best results of $68.1\%$ accuracy for FS action recognition on the SSv2 dataset.

\begin{figure}[t]
    \centering
    \includegraphics[clip=true, trim=2em 4em 0em 0em, width=1\columnwidth]{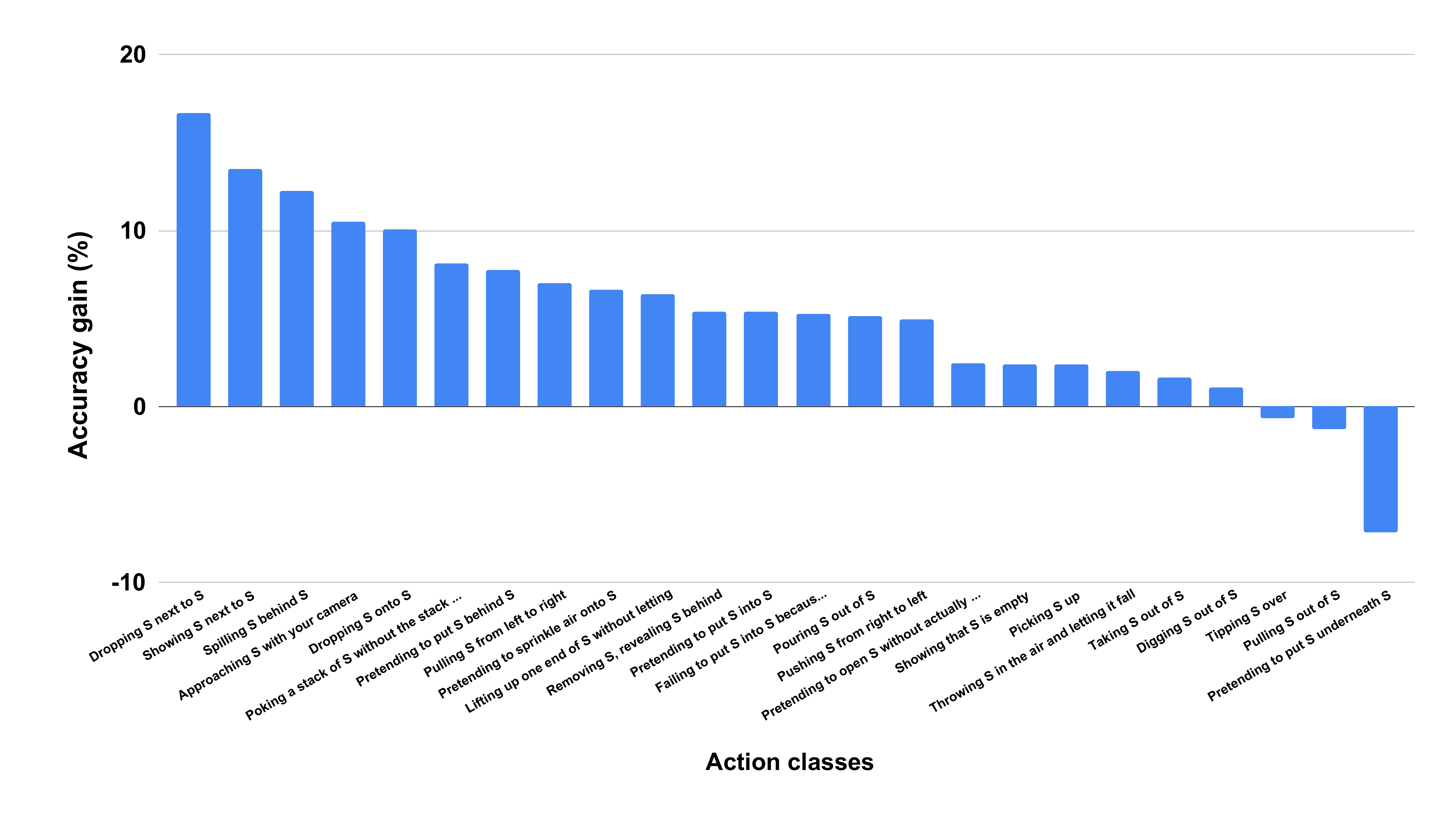}
    \vspace{-0.2cm}
    \caption{\textbf{Performance gains obtained by \texttt{STRM} over \texttt{Baseline TRM} on SSv2 test classes.} Our \texttt{STRM} achieves improved performance over \texttt{Baseline TRM} on $21$ out of $24$ test action classes in SSv2. Best viewed zoomed in.}
    \label{fig:per_class}
\end{figure}

\noindent\textbf{Class-wise performance gains:} Fig.~\ref{fig:per_class} shows the class-wise gains obtained by the proposed \texttt{STRM} framework over \texttt{Baseline TRM} on the SSv2 dataset. We observe that our \texttt{STRM} achieves gains above $10\%$ for classes such as \textit{Dropping something next to something}, \textit{Showing something next to something}, \etc. Out of $24$ action classes in the test set, our \texttt{STRM} achieves performance gains on $21$ classes. These results show that enriching the features by encoding the spatio-temporal contexts aids in improving the feature discriminability, leading to improved FS action recognition performance.

\section{Additional Qualitative Results\label{sec:qual}}
\begin{figure*}[t]
    \centering
    \includegraphics[width=\textwidth]{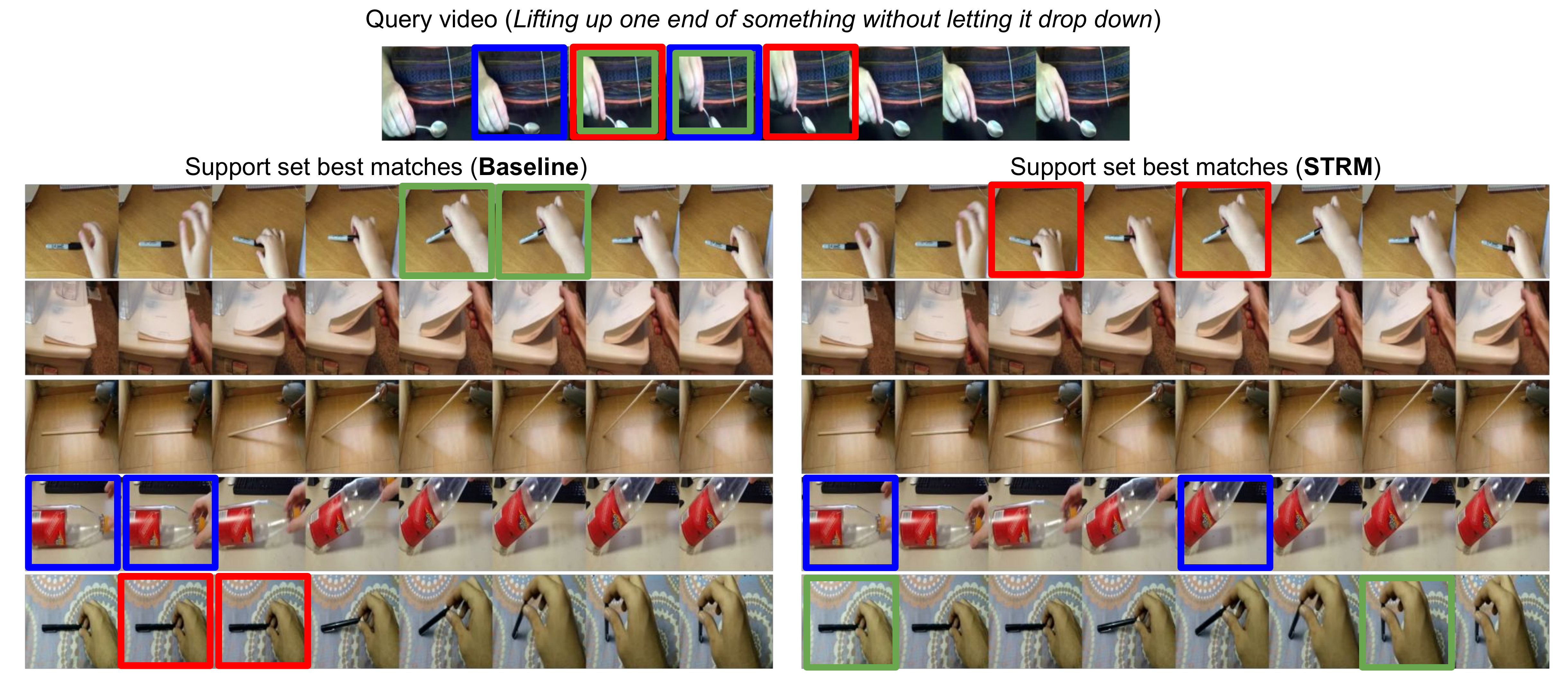}
    \vspace{-0.6cm}
    \caption{\textbf{Qualitative comparison between \texttt{Baseline TRM} and our \texttt{STRM} \wrt tuple matches.} Three query tuples of cardinality two are shown in red, green and blue for the query video at the top. Their corresponding best matches in the support videos (of ground-truth action) obtained by \texttt{Baseline TRM} and our \texttt{STRM} are shown on the left and right, respectively. The best matches for the blue and red tuples ($4^{th}$ and $5^{th}$ support videos) in \texttt{Baseline TRM} do not encode the action completely and are less discriminative.  We observe that our \texttt{STRM} is able to capture better matches with longer temporal variations (green and blue tuples in $4^{th}$ and $5^{th}$ support videos) due to the learned higher order temporal representations. See Sec.~\ref{sec:qual} for additional details.\vspace{1cm}}
    \label{fig:qual1}
\end{figure*}

\begin{figure*}[t]
    \centering
    \includegraphics[width=\textwidth]{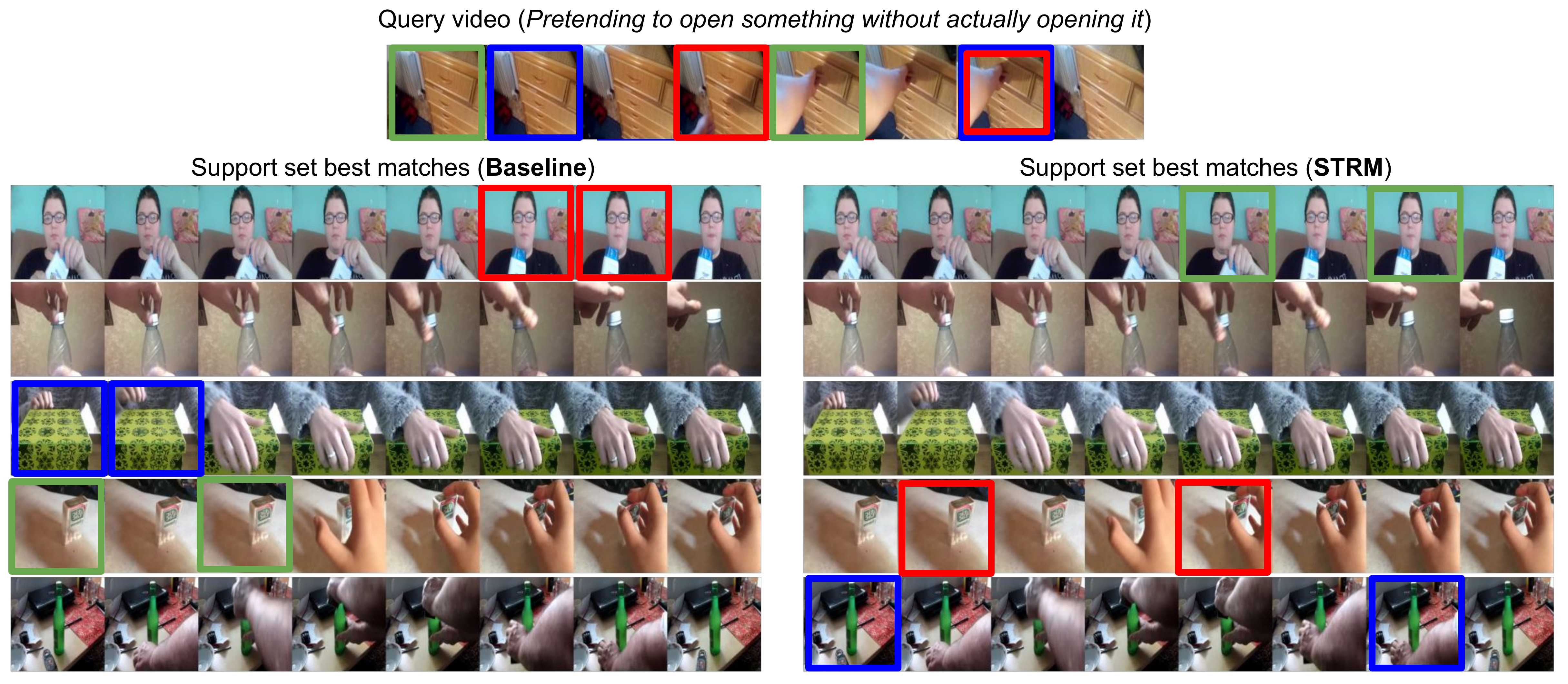} \vspace{-0.6cm}
    \caption{\textbf{Qualitative comparison between \texttt{Baseline TRM} and our \texttt{STRM} \wrt tuple matches.} See Fig.~\ref{fig:qual1} and Sec.~\ref{sec:qual} for additional details. The best matches for red and blue query tuples obtained by \texttt{STRM} ($4^{th}$ and $5^{th}$ support videos) are better representatives of the corresponding query tuples, in comparison to the best matches found by \texttt{Baseline TRM} ($1^{st}$ and $3^{rd}$ support videos).\vspace{1cm}}
    \label{fig:qual2}
\end{figure*}

\begin{figure*}[t]
    \centering
    \includegraphics[width=\textwidth]{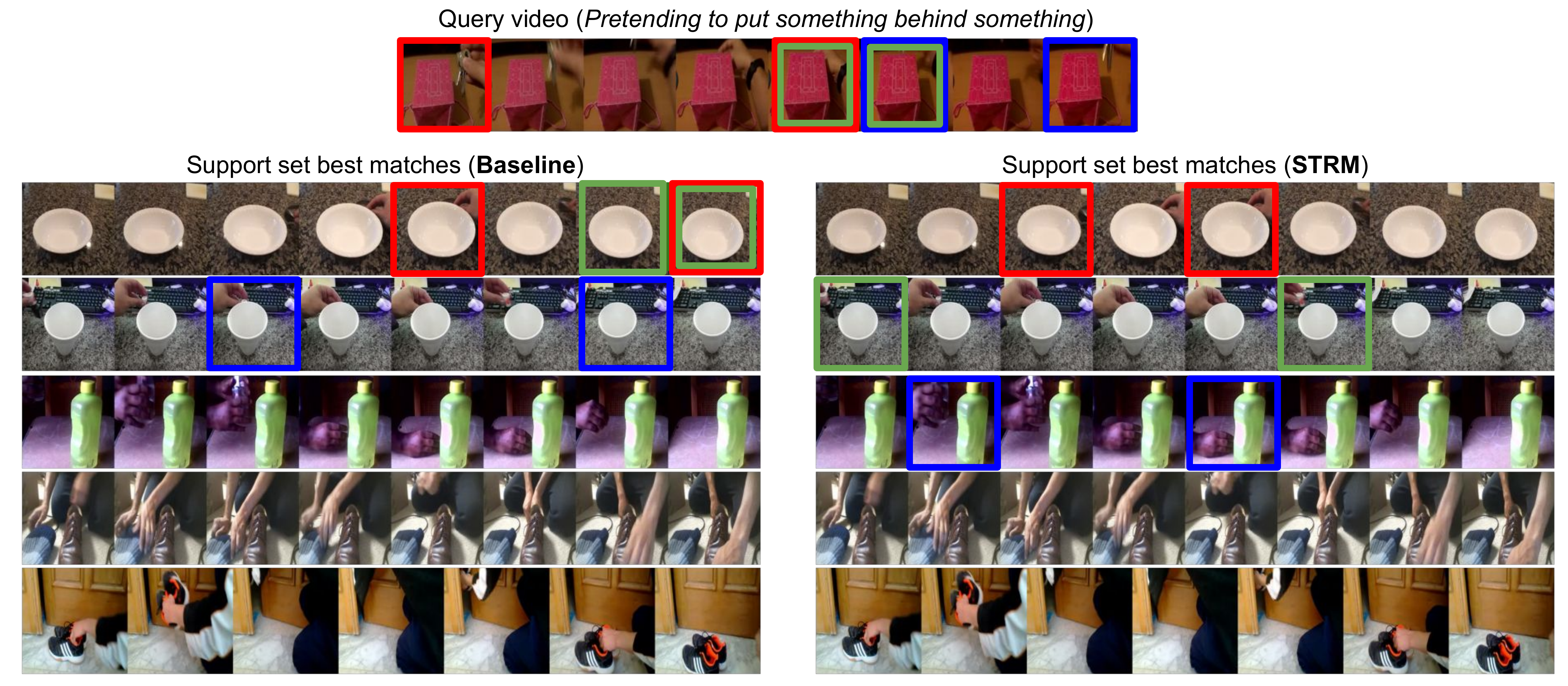}
    \caption{\textbf{Qualitative comparison between \texttt{Baseline TRM} and our \texttt{STRM} \wrt tuple matches.} See Fig.~\ref{fig:qual1} and Sec.~\ref{sec:qual} for additional details. For the query tuple in green, the best match obtained by our \texttt{STRM} ($2^{nd}$ support video) is a better representative, in comparison to the best match of \texttt{Baseline TRM} ($1^{st}$ support video).\vspace{1.5cm}}
    \label{fig:qual3}
\end{figure*}

\begin{figure*}[t]
    \centering
    \includegraphics[width=\textwidth]{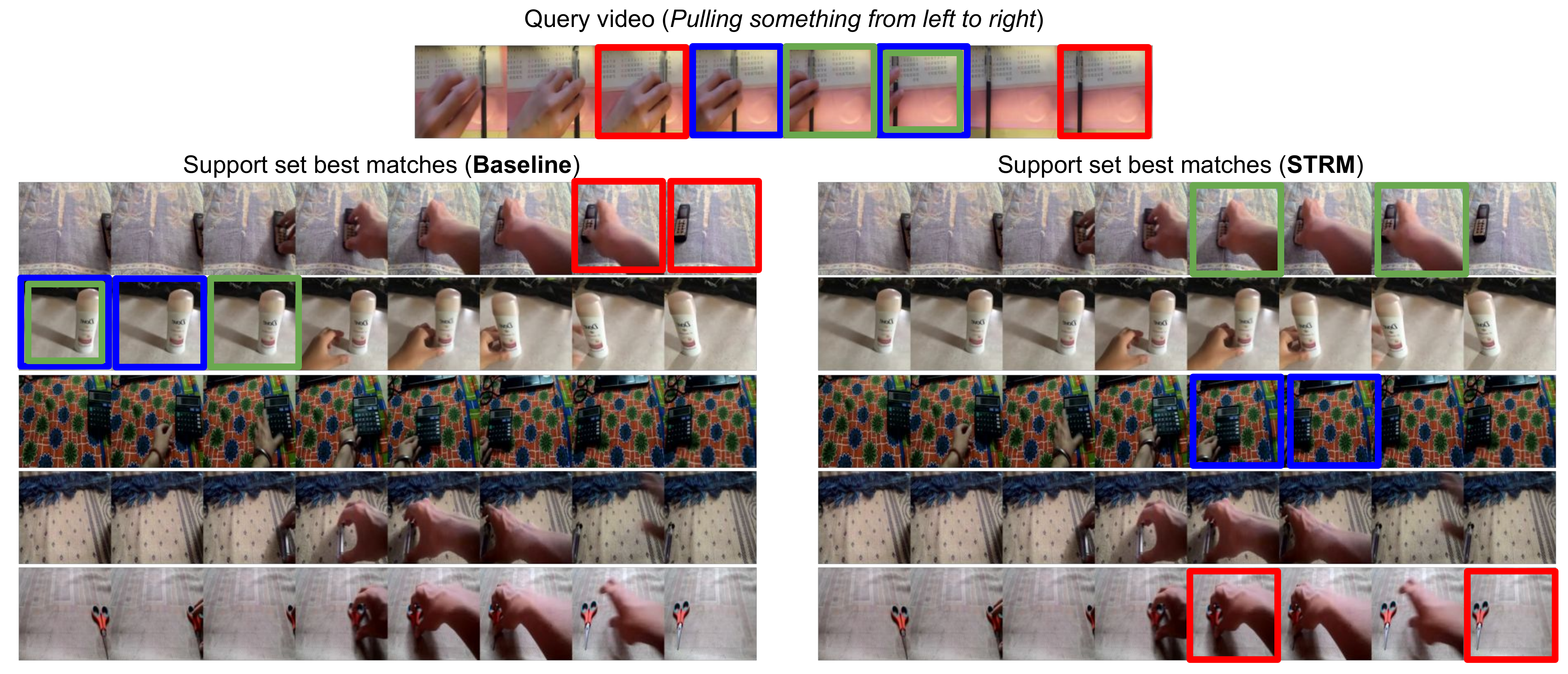}
    \caption{\textbf{Qualitative comparison between \texttt{Baseline TRM} and our \texttt{STRM} \wrt tuple matches.} See Fig.~\ref{fig:qual1} and Sec.~\ref{sec:qual} for additional details. The best matches found by \texttt{Baseline TRM} ($2^{nd}$ support video) for the green and blue query tuples fail to encode the true motion occurring in the corresponding query tuples. This is mitigated in the best matches obtained by our \texttt{STRM}.\vspace{1.5cm}}
    \label{fig:qual4}
\end{figure*}

\begin{figure*}[t]
    \centering
    \includegraphics[width=\textwidth]{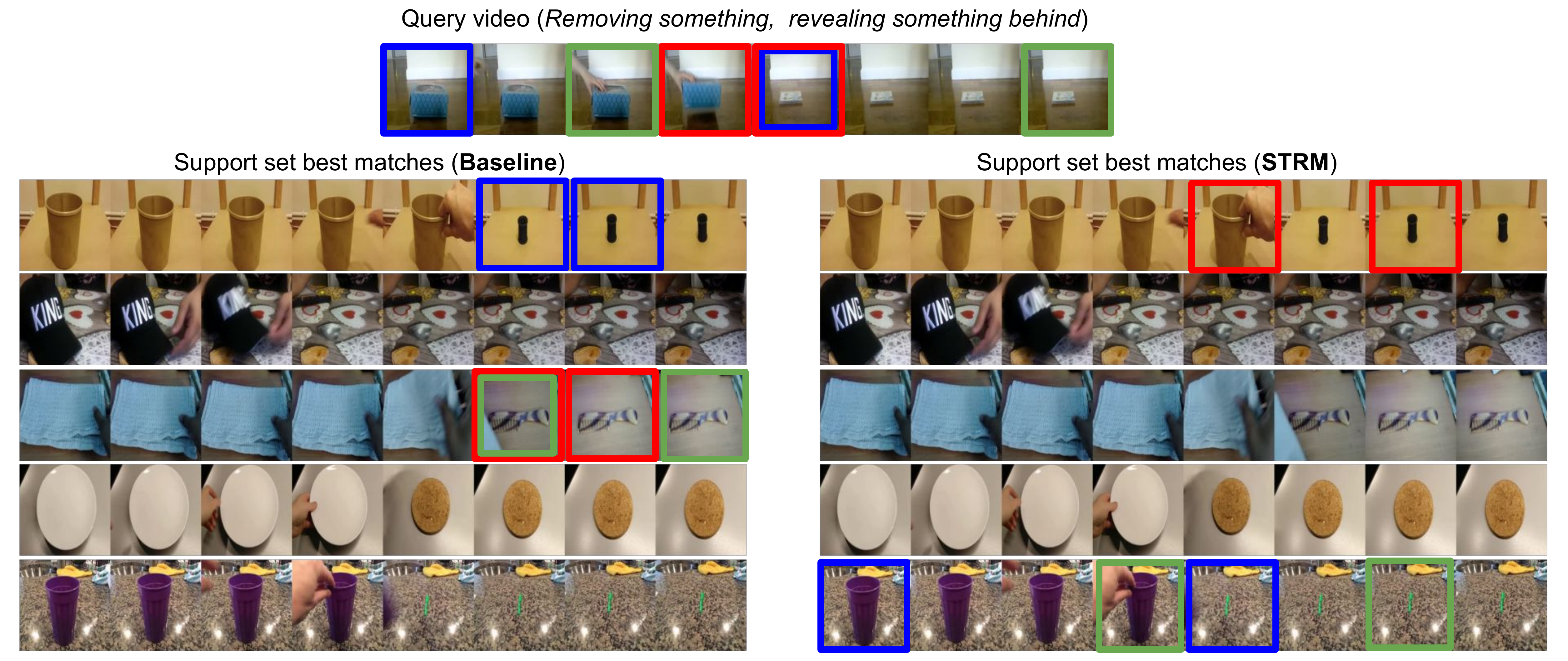}
    \caption{\textbf{Qualitative comparison between \texttt{Baseline TRM} and our \texttt{STRM} \wrt tuple matches.} See Fig.~\ref{fig:qual1} and Sec.~\ref{sec:qual} for additional details. The \texttt{Baseline TRM} fails to obtain support tuples that are representative enough for the query tuples in red and green. Our \texttt{STRM} alleviates this issue and obtains good representative matches ($1^{st}$ and $5^{th}$ support videos) since it enhances the feature disriminability through patch-level as well as frame-level enrichment and learns higher-order temporal representations.\vspace{2cm}}
    \label{fig:qual5}
\end{figure*}

\begin{figure*}[t]
    \centering
    \includegraphics[width=\textwidth]{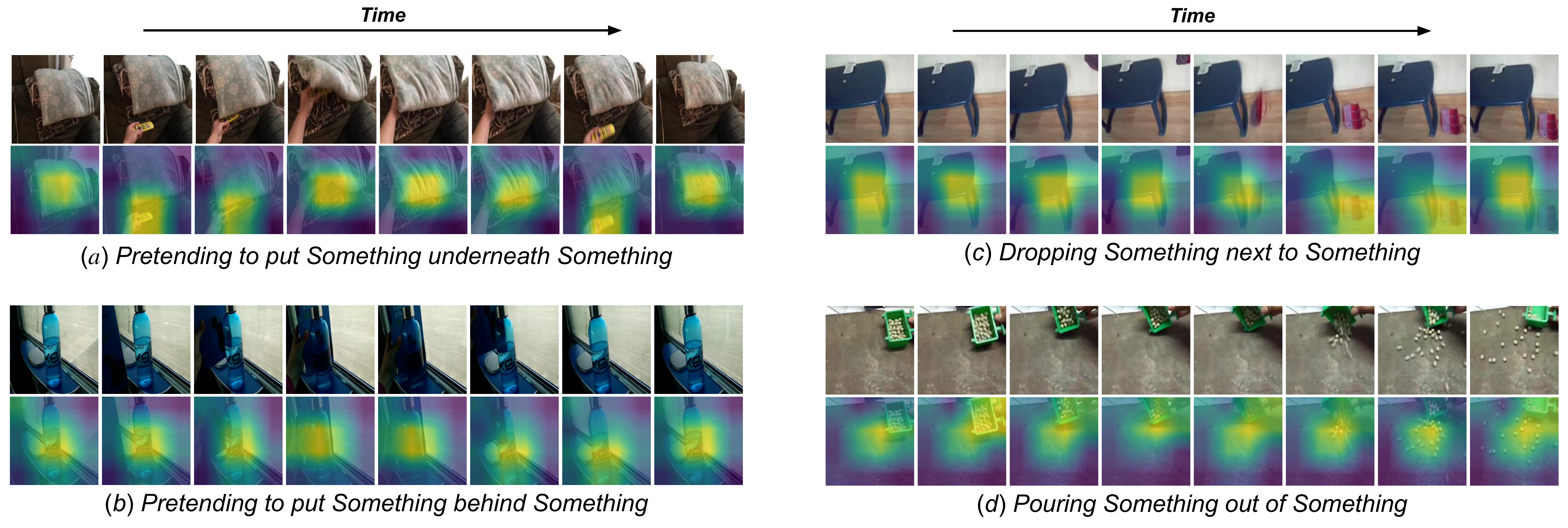}
    \caption{\textbf{Attention map visualization for four example classes in SSv2.} Our \texttt{STRM} learns to emphasize relevant spatio-temporal features that are discriminative, leading to improved FS action recognition performance. For instance, relevant objects for corresponding actions are emphasized: \textit{remote} in frames 2, 3 and 7 in (a), \textit{gems} in frames 4, 6, 7 and 8 in (d), respectively. Similarly, in (c), the focus on the the \textit{table} early on shifts to the \textit{bag}'s position in frames 6 and 7, which is required to reason out the \textit{Dropping Something next to Something} action.\vspace{2cm}}
    \label{fig:qual6}
\end{figure*}

Here, we present additional qualitative results \wrt tuple matching between query and support actions in Fig.~\ref{fig:qual1} to~\ref{fig:qual5}. In each example, a query video is shown on the top along with its ground-truth class name. Three query tuples of cardinality two are shown in red, green and blue. Their corresponding best matches in the support videos (of ground-truth action) obtained by \texttt{Baseline TRM} and our \texttt{STRM} are shown on the left and right, respectively. Generally, we observe that the best matches obtained by \texttt{Baseline TRM} do not encode the same representative features as in the corresponding query tuple. \Eg, blue and red tuples in $4^{th}$ and $5^{th}$ support videos of Fig.~\ref{fig:qual1}, red and blue tuples in $1^{st}$ and $3^{rd}$ support videos of Fig.~\ref{fig:qual2}. These results show that hand-crafted temporal representations in \texttt{Baseline TRM} are likely to not encode class-specific spatio-temporal context at lower cardinalities.
In contrast, our \texttt{STRM} obtains best matches that are highly representative of the corresponding query tuples and also encodes longer temporal variations. \Eg, green and blue tuples in $4^{th}$ and $5^{th}$ support videos of Fig.~\ref{fig:qual1}, blue tuple in $5^{th}$ support video of Fig.~\ref{fig:qual2}. The improved tuple matching between query and support actions in \texttt{STRM} is due to the proposed spatio-temporal feature enrichment, comprising patch-level and frame-level enrichment, which enhances the feature discriminability and the learning of the higher-order temporal representations at lower cardinalities that improves the model flexibility.
Furthermore, Fig.~\ref{fig:qual6} shows additional attention map visualizations on four example (novel) classes in the SSv2 dataset. Our \texttt{STRM} is able to emphasize the action-relevant objects in the video reasonably well. \Eg, in Fig.~\ref{fig:qual6}(a), \textit{remote} is emphasized in frames 2, 3 and 7. Similarly, while the \textit{bag}'s position is emphasized in frames 6 and 7, the focus is on the \textit{table} early on, which is required to reason out the \textit{Dropping Something next to Something} action in Fig.~\ref{fig:qual6}(c). We also observe that fine-grained novel actions with subtle 2D motion differences are harder to classify, \eg, \textit{Pretending to put Something behind Something} \vs \textit{Pretending to put Something underneath Something}. In general, our \texttt{STRM} learns to emphasize relevant spatio-temporal features that are discriminative, leading to improved FS action recognition performance.


In summary, the quantitative and qualitative results together emphasize the benefits of our proposed spatio-temporal enrichment module in enhancing feature discriminability and model flexibility, leading to improved few-shot action recognition.

\section{Additional Implementation Details}
The input videos are rescaled to a height of $256$ and
$L\neweq 8$ frames are uniformly sampled, as in~\cite{trx}. Random $224\times 224$ crops are used as augmentation during training. In contrast, only a centre crop is used during evaluation. We use the PyTorch~\cite{paszke2019pytorch} library to train our \texttt{STRM} framework on four NVIDIA 2080Ti GPUs. Since only a single few-shot task can fit in the memory, the gradients are accumulated and backpropagated once every $16$ iterations.


{\small
\bibliographystyle{ieee_fullname}
\bibliography{egbib}
}


\end{document}